\documentclass[journal]{IEEEtran}
\ifCLASSINFOpdf
  \usepackage[pdftex]{graphicx}
\else
\fi
\usepackage{amsmath}
\usepackage{subeqnarray}
\usepackage{amsfonts}
\usepackage{relsize}
\usepackage{booktabs}
\usepackage{multirow}
\usepackage{subfigure}
\usepackage{amssymb}
\usepackage{xcolor}
\usepackage[switch]{lineno}
\usepackage{hyperref}
\usepackage[ruled,vlined]{algorithm2e}
\usepackage{floatrow}

\hypersetup{
    colorlinks=true,
    urlcolor=blue,
}

\begin{document}
%
\title{AdaCrowd: Unlabeled Scene Adaptation for Crowd Counting}
%
%
%

\author{Mahesh Kumar~Krishna Reddy,  Mrigank~Rochan, Yiwei~Lu, Yang~Wang

\thanks{This work was done when all the authors were at the University of Manitoba.}
\thanks{M.K.K. Reddy is with the School of Computing Science, Simon Fraser University, Canada (Email: mahesh\_reddy@sfu.ca)}
\thanks{M. Rochan is with the Department of Computer Science, University of Manitoba, Canada (Email: mrochan@cs.umanitoba.ca)}
\thanks{Y. Lu is with the Department of Computer Science, University of Waterloo, Waterloo, Canada (Email: y485lu@uwaterloo.ca)}
\thanks{Y. Wang is with the Department of Computer Science, University of Manitoba and Huawei Technologies Canada (Email: ywang@cs.umanitoba.ca)}
}

%
%

\markboth{Journal of \LaTeX\ Class Files,~Vol., No., ~2020}
{Shell \MakeLowercase{\textit{et al.}}: Bare Demo of IEEEtran.cls for IEEE Journals}
%



\maketitle

\begin{abstract}
We address the problem of image-based crowd counting. In particular, we propose a new problem called \emph{unlabeled scene-adaptive crowd counting}. Given a new target scene, we would like to have a crowd counting model specifically adapted to this particular scene based on the target data that capture some information about the new scene. In this paper, we propose to use one or more unlabeled images from the target scene to perform the adaptation. In comparison with the existing problem setups (e.g. fully supervised), our proposed problem setup is closer to the real-world applications of crowd counting systems. We introduce a novel {\fontfamily{qcr}\selectfont AdaCrowd} framework to solve this problem. Our framework consists of a crowd counting network and a guiding network. The guiding network predicts some parameters in the crowd counting network based on the unlabeled images from a particular scene. This allows our model to adapt to different target scenes. The experimental results on several challenging benchmark datasets demonstrate the effectiveness of our proposed approach compared with other alternative methods.  Code is available at {\emph{\url{https://github.com/maheshkkumar/adacrowd}}}
\end{abstract}


%
\IEEEpeerreviewmaketitle


\section{Introduction}
\IEEEPARstart{I}n recent years, image-based crowd counting has become an active area of research due to its potential applications in numerous real-world domains, such as traffic monitoring, smart city planning, surveillance, security, etc. Given an input image, the goal of crowd counting is to estimate the number of people in the image. The most recent work in this area formulates the problem of estimating a density map for the input image. The pixel values in the density map indicate the crowd densities at different locations in the image. The final crowd count can be obtained by summing the values in the estimated density map. The most recent work uses various forms of convolutional neural networks (CNN) trained in a supervised manner to perform the density estimation given an image. There has been lots of effort towards designing effective CNN architectures~\cite{kang2018crowd,li2018csrnet,sam2019almost,sam2017switching,sindagi2017cnn,sindagi2017generating,walach2016learning,wang2019learning,yu2016multi,zhang15cross,zhang2016single} with increasing capacities to address the problem of crowd counting. However, these approaches require a large number of training images which are expensive to collect. This becomes more problematic if we want to train a crowd counting system tuned to a specific environment (which is the case of many surveillance applications) due to the burden of collecting a large number of training images from the target environment.

\begin{figure}[!t]
  \centering
  \includegraphics[height=6cm, width=8cm]{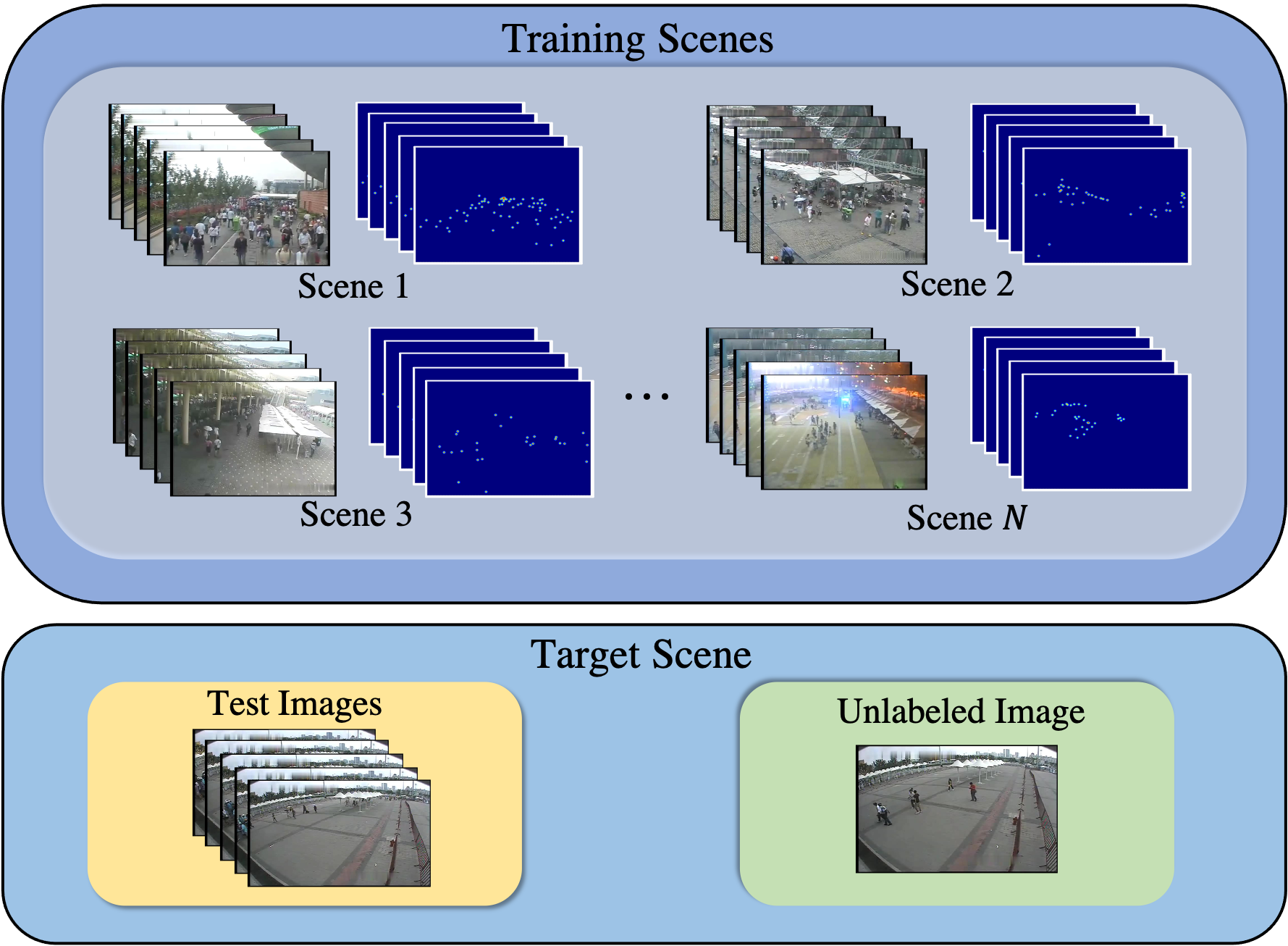}
  \caption{Illustration of the unlabeled scene-adaptive crowd counting problem. (Top row) Our training data consists of images and labels collected from $N$ different scenes. (Bottom row) During testing, we are given one or more unlabeled images from a new target scene. Our goal is to produce a crowd counting model specifically adapted to this target scene to predict crowd count on the test images from this scene. Unlike the few-shot adaptation in \cite{hossain2019one,Reddy_2020_WACV}, our problem setup does not require any labeled images from the target scene for adaptation.}
  \label{fig:problem_setup}
\end{figure}

Some recent works~\cite{hossain2019one,Reddy_2020_WACV} propose to address few-shot scene-adaptive crowd counting. The problem setup in \cite{hossain2019one,Reddy_2020_WACV} is motivated by the following observation. In many real-world applications, the crowd counting model only needs to work well in a very specific scene. Let us consider the application of video surveillance. Since the surveillance camera is fixed at a particular location, all test images belong to the same scene. As a result, we only need the crowd counting model to work well in this particular camera scene. The problem setup in \cite{hossain2019one,Reddy_2020_WACV} assumes that we have access to a small number (e.g., one to five) of labeled images from the target scene where the model will be deployed. The model is then learned to effectively adapt to a new target scene using these few labeled examples.

Compared with the standard supervised setting, the few-shot scene adaptation setup in \cite{hossain2019one,Reddy_2020_WACV} brings crowd counting closer to real-world deployment. However, this setup still has some limitations. First of all, it still requires at least one labeled image from the end-user. Although it has drastically reduced the data requirement compared with the supervised case, it is still a burden for the end-user.  In particular, for crowd counting applications, labeling an image requires annotating the location of each person in the image. Since there can be hundreds of persons in an image, the required labeling effort is non-trivial from the end-user's viewpoint.  Second, when deploying to the target scene, the scene adaptation method in \cite{hossain2019one,Reddy_2020_WACV} involves fine-tuning several layers of a CNN model. This requires running gradient updates and backpropagation through the network for several iterations. In practice, the computation requirement is still too high for typical surveillance cameras with limited computing capabilities.  In some deployment environments (e.g., Tensorflow Lite), it may not be possible to run the backpropagation to compute the gradients. 

Inspired by \cite{hossain2019one,Reddy_2020_WACV}, we push the envelope even further by proposing a new problem called \emph{Unlabeled Scene-Adaptive Crowd Counting}. Similar to \cite{hossain2019one,Reddy_2020_WACV}, our objective is to have a model adapted to a specific target scene during deployment. But different from \cite{hossain2019one,Reddy_2020_WACV}, we do not require any labeled images from the target scene. Our adaptation method only requires one or more \emph{unlabeled} images from the target scene for adaptation (see Fig.~\ref{fig:problem_setup}). Since unlabeled images are fairly easy to collect in practice, this problem setup greatly reduces the data annotation effort from the user. Besides, our proposed approach also significantly reduces the required computation during adaptation. It only involves some feed-forward computation without backpropagation.
 
We call our model {\fontfamily{qcr}\selectfont AdaCrowd}. It consists of two neural networks: a \textit{crowd counting network} and a \textit{guiding network}. The crowd counting network has some parameters that will be adapted to each scene. The guiding network is learned to map the unlabeled images from a scene to the adaptable parameters in the crowd counting network. The parameters of these two networks are learned in a way that allows effective adaptation to a new scene using only a few unlabeled images from that scene.

\noindent\textbf{Summary of Contributions:} The contributions of this work are manifold. First, we present a new problem called unlabeled scene-adaptive crowd counting. Different from the few-shot adaptation setup in \cite{hossain2019one,Reddy_2020_WACV}, our problem formulation only uses unlabeled images from the target scene for adaptation. Second, we develop an approach termed as {\fontfamily{qcr}\selectfont AdaCrowd} for learning the model parameters. Our model can effectively adapt to a new scene given the unlabeled images from that scene without fine-tuning at test time. Finally, we conduct an extensive evaluation of the proposed approach on several challenging benchmark datasets. Our approach significantly outperforms other alternatives.

\section{Related Work}

In this section, we review prior work in two main lines of research most relevant to our work, namely crowd counting and few-shot learning.

\noindent{\bf Crowd Counting:} Some early work uses regression-based approaches. The traditional methods resort to hand-crafted features (such as LBP, HOG, GLCM) to train various regression models (such as linear~\cite{paragios2001mrf}, piecewise linear~\cite{chan2008privacy,chen2012feature} and Gaussian process regression~\cite{marana1998efficacy}) to predict the crowd count. More recently, some work~\cite{chattopadhyay2017counting,shang2016end,wang2015deep} employs deep CNNs for end-to-end crowd count estimation. 

A lot of recent work uses density-based methods that learn to estimate a density map of an input image. Lempitsky et al.~\cite{lempitsky2010learning} learn a linear mapping function from the local image patches to their corresponding density representation, where an integral over the density map results in the crowd count. Pham et al.~\cite{pham2015count} propose to learn a non-linear mapping function by employing a random forest regression on multiple local image patches based on the voting mechanism for their corresponding densities to address the large variation in crowd images. Some follow up works~\cite{cheng2019learning,li2018csrnet,ma2019bayesian,sam2019almost,sam2017switching,sindagi2017generating,wan2019adaptive,wang2019learning,yu2016multi,zhang15cross,zhang2016single,liu2020adaptive,jiang2020density,liu2020denet,gao2020nwpu,jiang2020learning,gao2020pccnet,Jiang_2020_CVPR} exploit the non-linearity of convolutional neural networks to learn the crowd density maps on either image patches or on the whole images. Early CNN-based methods exploit multi-column architecture~\cite{kang2018crowd,sindagi2017cnn,sindagi2017generating,walach2016learning,zhang2016single} with different receptive fields to learn features at diverse scales. Sam et al.~\cite{sam2017switching} propose a method to choose the best regression network with a specific receptive field for an image patch based on the label prediction by a classifier. Li et al.~\cite{li2018csrnet} use dilated convolutional layers to expand the receptive field in place of pooling layers to address scale variations. 

There has been work that aims to address the domain shift between training and test images in crowd counting. Wang et al.~\cite{wang2019learning} address the limitation of labeled real-world data by using domain adaptation from synthetic images to real images. However, this method requires prior knowledge of the crowd density statistics in the target scene to carefully select matching images in the synthetic data to overcome negative adaptation. Hossain et al.~\cite{hossain2019one} propose an approach of adaptation from a source to target scene using one labeled image. Reddy et al.~\cite{Reddy_2020_WACV} propose a few-shot based meta-learning approach to perform scene-adaptive crowd counting.  Wang et al.~\cite{wang2021neuron} address domain shift by exploiting the parameter difference between the source and the target models. More specifically, a source model is trained on the synthetic source dataset. Given a real-world target dataset with few labeled images, they learn scale and shift parameters for each neuron of the source model to generate a target-specific model through a linear transformation. However, these approaches~\cite{hossain2019one,Reddy_2020_WACV,wang2021neuron} still require labeled data from the target scene to adapt the model. In contrast, our approach does not require any labeled data from the target dataset during adaptation.

Kang et al.~\cite{kang2017incorporating} propose to use some meta-information about the target scene (e.g., camera tilt angle, height, or perspective map) to adapt the weights in convolutional layers. In practice, the meta-information is not always available. Most benchmark datasets in crowd counting do not provide this meta information. So this limits the applicability of this method.  In contrast, the unlabeled scene adaptation setup proposed in our work requires minimal effort in terms of data collection from the end-users and can be broadly applied in practical scenarios.  

\noindent{\bf Few-Shot Learning:} Few-shot learning, in general, has been widely studied in the context of image classification for learning novel classes with few data points. Some early work~\cite{fei2006one,salakhutdinov2012one} uses a Bayesian method to transfer the prior knowledge about the appearance from learned categories to novel categories. Recently, meta-learning~\cite{finn2017model,koch2015siamese,nichol2018first,snell2017prototypical,sung18learning,vinyals2016matching} has been explored for the quick adaptation to new categories. More specifically, the metric-based meta-learning approaches~\cite{koch2015siamese,munkhdalai2017meta,santoro2016meta,snell2017prototypical,sung18learning,vinyals2016matching} learn the metric as a kernel function to measure the similarity between embedding vectors of two data points. In model-based meta-learning methods~\cite{munkhdalai2017meta,santoro2016meta}, the goal is to update the model's parameters for fast learning. The parameter update can be accomplished either from an internal architecture~\cite{santoro2016meta} or by prediction from another network~\cite{munkhdalai2017meta}. In contrast, the optimization-based meta-learning models ~\cite{finn2017model,nichol2018first,ravi2016optimization} tweak the gradient-based optimization to learn from few examples and converge within a small number of gradient steps.

\section{Problem Setup}\label{sec:setup}
In this section, we first review three standard setups for crowd counting that have been studied in the literature. The various limitations of these setups in real-world deployment motivate us to introduce a new problem setup for crowd counting. We believe our setup is closer to real-world scenarios compared with these existing problem setups.

\noindent{\bf Supervised:} This is the most common setup in previous work. This setup treats crowd counting as a purely supervised learning problem. The goal is to learn a function $f_{\theta}:x \rightarrow y$ that maps an image $x$ to a density map $y$. The model parameters $\theta$ are learned from labeled training images. There has been lots of previous work on designing powerful models~\cite{cheng2019learning,kang2018crowd,li2018csrnet,ma2019bayesian,sam2019almost,sam2017switching,sindagi2017cnn,sindagi2017generating,walach2016learning,wan2019adaptive,wang2019learning,yu2016multi,zhang15cross,zhang2016single} in this supervised setting.

\noindent{\bf Domain Adaptation (DA):} The standard supervised approach implicitly assumes that training and test images are similar. In practice, training and test images often come from different domains, e.g. they might be collected from two different scenes. Due to the domain shift, a model trained on the source domain often does not perform well in the target domain. Domain adaptation is a standard approach to address this domain shift. The most recent work focuses on unsupervised domain adaptation (UDA). In UDA, the source domain contains labeled data, whereas the target domain only contains unlabeled data. Most approaches of UDA~\cite{long2015learning,long2017deep} use a domain adaptation loss to minimize the discrepancy between features of source and target domains. 

The domain adaptation setup also has limitations in practical deployment. First of all, DA assumes that we have enough unlabeled data from the target domain. This might be infeasible in practice. For instance, if we consider an end-user's environment as the target domain, we may not have the authority to collect images from the target domain. Second, even if we have enough unlabeled data in the target domain, most DA approaches still require running an algorithm to perform many iterations of gradient updates and backpropagation. In practice, a crowd counting system might be deployed directly on end-users' surveillance cameras or other devices that may not have enough computing capabilities to run the domain adaptation algorithm.

\noindent {\bf Few-Shot Scene Adaptation:} Some recent works \cite{hossain2019one,Reddy_2020_WACV} introduce a new problem setup called few-shot scene adaptation. During deployment, this setup only needs a small number (e.g. one to five) of \emph{labeled} images from a target scene. The training data consists of labeled images from multiple scenes. The model is learned in a way that enables it to quickly adapt to a new scene with only a few labeled examples. These works argue that it is often easier to get a small number of images (even with labels) in the target scene. For example, after a surveillance camera is installed, there is often a calibration process. It is possible to collect (and even label) a few images during this calibration process.
Although this setup brings us closer to real-world scenarios, it still requires a few labeled images from the target domain (or scene). Additionally, the methods in \cite{hossain2019one,Reddy_2020_WACV} involve several rounds of gradient updates and backpropagation to adapt model parameters based on the few labeled images. In practice, this is still taxing for end-users. In this paper, we push the limit on data requirements and overcome gradient updates at test time by proposing the following setup.

\noindent {\bf Unlabeled Scene Adaptation (this paper):} This setup is similar to the few-shot scene adaptation. The key difference is that this setup does not require labeled examples from the target scene during deployment. Instead, it only requires a small number of unlabeled images from the target scene as they are fairly easy to collect in practice. Similar to \cite{hossain2019one,Reddy_2020_WACV}, our training data consist of labeled images from multiple scenes. We propose a novel approach that learns to adapt to a target scene using only the scene-specific unlabeled images. Our approach requires minimal data collection effort from end-users. It only involves some feed-forward computation (i.e., no gradient update or backpropagation) for adaptation. This makes it more practical and suitable for real-world deployment.

\section{Our Approach}\label{sec:approach}

\begin{figure*}[ht]
	\centering
	\includegraphics[width=18cm,height=5cm]{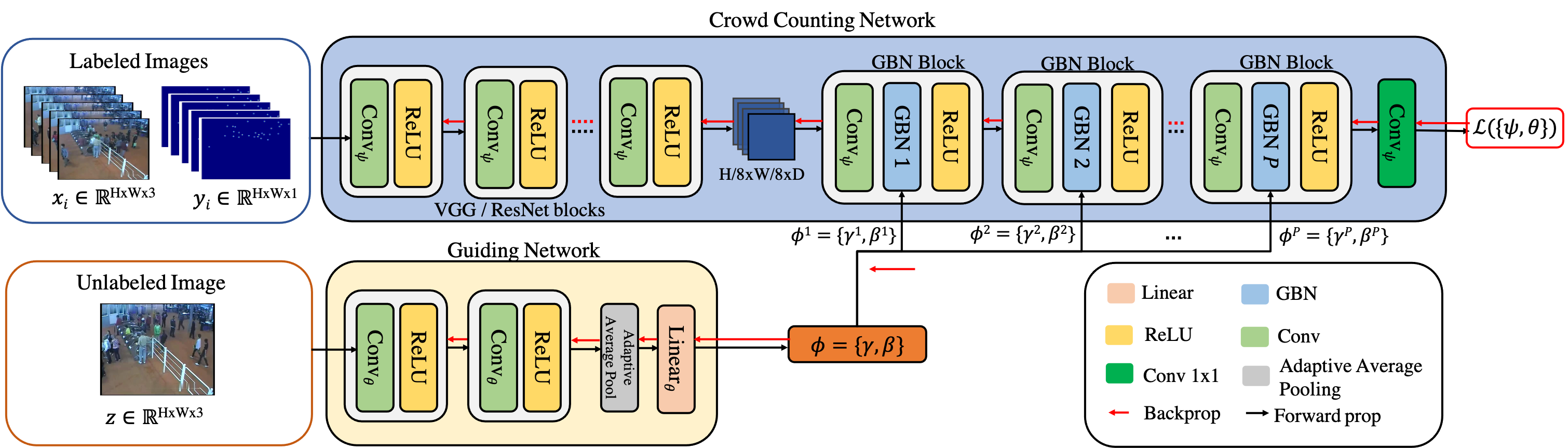}
	\caption{Illustration of our  {\fontfamily{qcr}\selectfont AdaCrowd} framework. (Top row) The crowd counting network takes an input image and estimates the density map. The crowd counting network has $P$ number of GBN blocks with the proposed guided batch normalization (GBN) layers. The parameters $\phi = \{\gamma, \beta\}$ for all the GBN layers are predicted from another neural network called the guiding network. (Bottom row) The guiding network takes the unlabeled image ($z$) from the training scene and produces GBN parameters specifically adapted to that scene. Through these scene-adapted GBN parameters, our model achieves adaptation to different scenes.}
	\label{fig:architecture_overview}
\end{figure*}

In Fig.~\ref{fig:architecture_overview}, we provide an overview of our {\fontfamily{qcr}\selectfont AdaCrowd} framework. The framework consists of two main networks, the \textit{crowd counting network} and the \textit{guiding network}. The crowd counting network is a CNN model that takes an input image and outputs its corresponding density map. Compared with standard crowd counting models, the key difference of our crowd counting network is that it has several special layers called the \emph{guided batch normalization (GBN)} layers. A GBN layer plays a role similar to the standard batch normalization (BN). The main difference is that the affine parameters of a BN layer are determined from the mini-batch of data, while the parameters of the GBN layer are directly predicted by the guiding network. Most parameters of the crowd counting network are shared across different scenes. But the parameters of GBN layers change to adapt to different scenes. Due to this adjustable nature of GBN parameters, our model can learn to adapt to different scenes. Another important component in the architecture is the guiding network. This network takes the unlabeled images from a specific target scene as its input and outputs the GBN parameters for this scene. During training, the guiding network learns to predict GBN parameters that work well for the corresponding scene. At test time, we use the guiding network to adapt the crowd counting network to a specific target scene.

\noindent{\bf Guided Batch Normalization:} In our {\fontfamily{qcr}\selectfont AdaCrowd} framework, we propose a new conditional normalization layer called the \emph{Guided Batch Normalization~(GBN)} layer. For ease of presentation, let us consider one particular layer in a CNN model and see how to design the GBN layer to be inserted after this layer in the model. We consider a mini-batch of $B$ examples ${x_{i}: i=1,2,...,B}$, where $x_i$ is the CNN feature map at this layer for the $i$-th example, i.e., $x_i \in \mathbb{R}^{H \times W \times D}$ where $H\times W$ are the spatial dimensions and $D$ is the channel dimension. Similar to batch normalization~(BN)~\cite{ioffe2015batch}, in GBN we normalize the activation to have zero mean and unit variance along the channel dimension over the mini-batch during training. However, unlike BN which learns the affine transformation parameters $\gamma$ and $\beta$ over the examples $\{x_i:i=1,2,...,B\}$ in the mini-batch, in GBN, we directly predict them using the guiding net. The overall computation in GBN can be summarized as follows:
\begin{subeqnarray}
  \label{eq:gbn_compute}
&&\hspace{-30pt} \hat{x}_i[h,w,d] = GBN(x_i[h,w,d];\gamma_{d},\beta_{d}), \textrm{ where}\label{eq:gbn_overview}\\
  &&\hspace{-30pt} GBN = \gamma_{d}\cdot \big( (x_i[h,w,d] - \mu_{d})/\sigma_{d} \big) + \beta_{d} \label{eq:gbn_complete}\\
&& \hspace{-30pt} \forall \ \ i\in\{1..B\}, h\in\{1..H\}, w\in\{1..W\}, d\in\{1..D\}
\end{subeqnarray}
where $x_i[h,w,d]$ denotes the $[h,w,d]$ entry of the feature map $x_i$ and $\hat{x}_i[h,w,d]$ is the normalized output of the activation $x_i[h,w,d]$. The mean $\mu_{d}$ and standard deviation $\sigma_{d}$ are calculated from the activation in channel $d$:
\begin{subeqnarray}
&&\hspace{-40pt}\mu_{d}= \frac{\sum_{i=1}^{B}\sum_{h=1}^{H}\sum_{w=1}^{W}x_i[h,w,d]}{B\times H \times W}\\
&&\hspace{-40pt}\sigma_{d} = \sqrt{\frac{\sum_{i=1}^{B}\sum_{h=1}^{H}\sum_{w=1}^{W}\big(x_i[h,w,d] - \mu_{d}\big)^{2}}{B\times H\times W} + \epsilon} 
\end{subeqnarray}                         
where $\mu_{d}, \sigma_{d} \in \mathbb{R}$ and $\epsilon$ is used for numerical stability. In Eq.~\ref{eq:gbn_compute}, the affine parameters $\gamma_d, \beta_d \in \mathbb{R}$ control the scaling and shifting operations corresponding to the $d$-th channel dimension. Let us use $\gamma$ and $\beta$ to denote the concatenations of $\{\gamma_d: \forall d\}$ and $\{\beta_d: \forall d\}$, respectively. The variables $\gamma, \beta\in\mathbb{R}^{D}$ are the parameters of this GBN layer and their values vary across different scenes. In general, for the $p$-$th$ GBN layer, we can denote the affine parameters for the feature channel $d$ by $\gamma^{p}_{d}, \beta^{p}_{d}$. 

\begin{figure}[h]
	\centering
	\includegraphics[height=6cm]{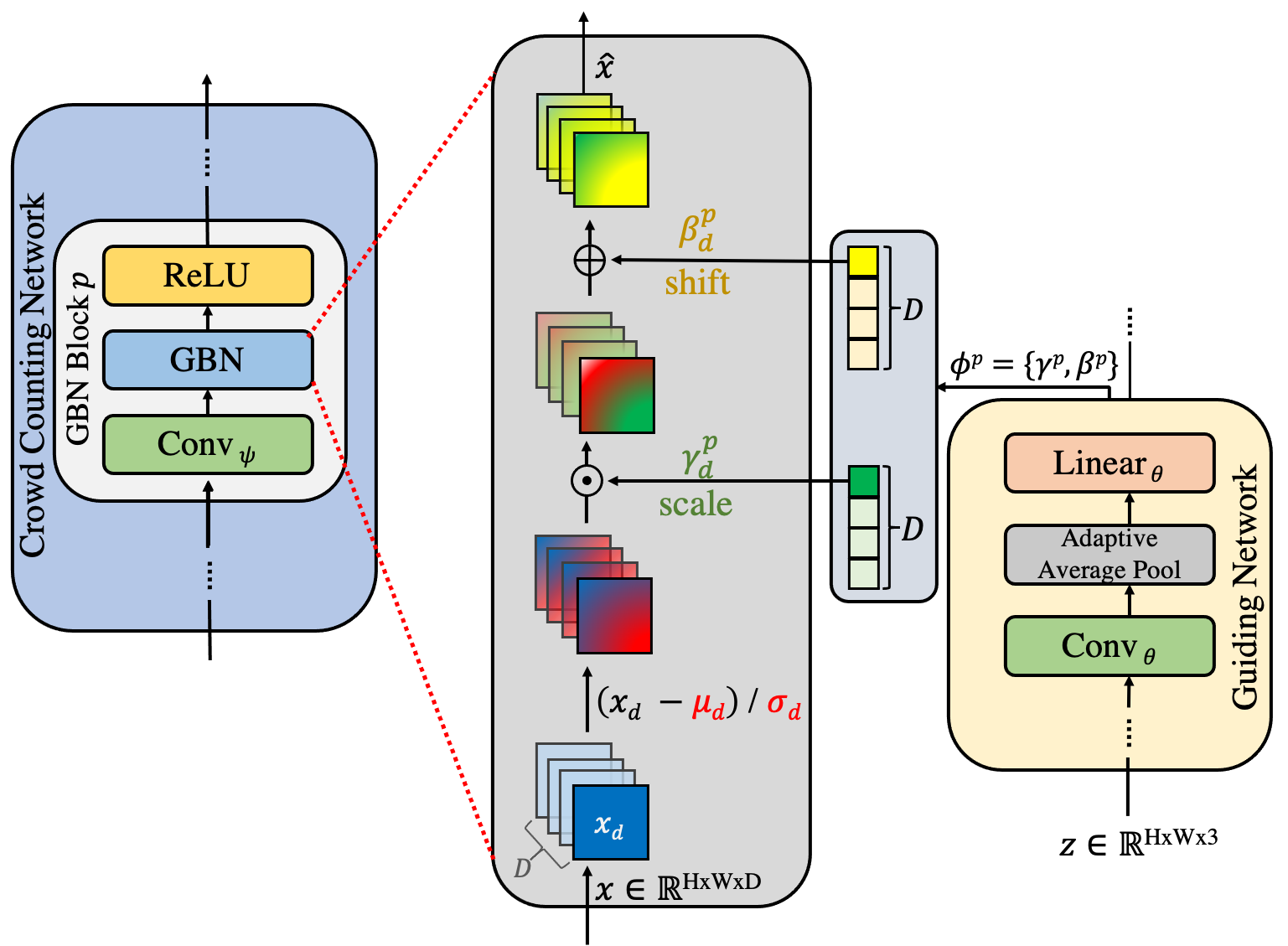}
	\caption{Overview of a GBN block in the {\fontfamily{qcr}\selectfont AdaCrowd} framework. For instance, given an input $x$ to the GBN layer in the GBN block $p$, we first normalize $x$ along the channel dimensions in the GBN layer. We then use the affine transformation parameters $\gamma^{p}_{d}$ and $\beta^{p}_{d}$ corresponding to the $p$-th GBN block to uniformly scale and shift the activation features to generate the output $\hat{x}$. The affine parameters are generated by the guiding network based on the scene-specific unlabeled data $z$.}
	\label{fig:gbn_block}
\end{figure}

The proposed GBN layer is related to conditional batch normalization (CBN)~\cite{de2017modulating}. However, the noticeable differences between them are: (1) CBN layer is specially devised to consider information from linguistic data, whereas GBN is designed to use information from spatial input data such as images. (2) CBN depends on the learned weights for $\gamma$ and $\beta$ obtained from BN layer in a pre-trained network for initialization. However, in GBN we directly predict $\gamma$ and $\beta$ from external visual data. As a result, we do not require a pre-trained network for GBN weight initialization. (3) Another technical difference is that, CBN learns to shift the pre-trained $\gamma$ and $\beta$ by small margins, whereas we completely replace the values for $\gamma$ and $\beta$ as each scene is visually different. Therefore, GBN is more intuitive and flexible when applying affine transformation on spatial data. We visualize the operations in a GBN block in Fig.~\ref{fig:gbn_block}.

\noindent {\bf Crowd Counting Network:} The objective of the crowd counting network is to generate the density map for the input image. For our {\fontfamily{qcr}\selectfont AdaCrowd} framework, we can use any backbone crowd counting network by inserting several GBN layers into the backbone. 

To simplify the notation, we use $\phi$ to denote the concatenation of all parameters in GBN layers (i.e., $\gamma$ and $\beta$ from all GBN layers) in the crowd counting network. 
We use $\psi$ to denote other parameters in the crowd counting network. The crowd counting network can be written as a function $f$ parameterized by $\{\phi,\psi\}$ as:
\begin{equation}
  \hat{y}=f(x;\{\phi, \psi\})
\end{equation}
where $f$ maps the input image $x$ to the predicted density map $\hat{y}$.

Note that $\psi$ will be the same for different scenes, while $\phi$ will change across scenes since $\phi$ is the output of the guiding network (described below). By predicting $\phi$ specifically to a scene, we achieve scene-adaptive crowd counting.


\noindent {\bf Guiding Network:} The goal of the guiding network is to predict the GBN parameters $\phi$ using one or more unlabeled images from a scene. For brevity, we assume that we have only one unlabeled image (denoted by $z \in \mathbb{R}^{H \times W \times 3}$) in the following and later we will describe a more generalized case to handle multiple unlabeled images.

Given the unlabeled image $z$, we use it as the input to the guiding network to predict the GBN parameters $\phi$ as: 
\begin{equation}
  \label{eq:gbn}
  \phi=g(z;\theta)
\end{equation}
where $g(\cdot;\theta)$ denotes a function parameterized by $\theta$. In this paper, $g(\cdot;\theta)$ is implemented as a CNN where the input $z$ is an unlabeled image. But in general, $g(\cdot;\theta)$ can be any arbitrary parametric function.

Our framework can be easily extended to the more general case where we have $K$ ($K>1$) unlabeled images from the target scene. In this case, We can simply average ($\phi = \frac{1}{K}\sum_{i=1}^{K}\phi_{i}$) the predicted $\phi_{i}$ over $K$ inputs.

Note that the architectures used for the encoder part of crowd counting network and guiding network do not share any parameters as the objectives for these networks are different.  The crowd counting encoder is targeted towards extracting scene-invariant features, so it can generalize across different scenes. In comparison, the guiding network learns to extract scene-variant information that helps unlabeled scene adaptation.

\noindent {\bf Learning and Inference:} Note that the counting network is parameterized by $\{\phi, \psi\}$. But the model parameters we need to learn are $\{\psi, \theta\}$, since $\phi$ will be directly predicted from the guiding network (see Eq.~\ref{eq:gbn}). We would like to learn model parameters that have the following property. Suppose we have a new scene represented by $z$ (unlabeled image). We would like the counting network with GBN parameters $\phi$ obtained via Eq.~\ref{eq:gbn} to perform well on images from this scene. To achieve this, we learn the model parameters from a set of labeled training images collected from multiple scenes. The model parameters are learned in a way that is amenable to effective adaptation to a new scene based on its scene-specific unlabeled data $z$.

The learning algorithm works iteratively. In each iteration, the algorithm constructs a unlabeled scene adaptation task that mimics the scenario during testing. For a scene $\mathcal{S}^{t}$, we use $\{(x_1,y_1),...,(x_N,y_N)\}$ to denote the $N$ training examples from this scene, where $(x_i,y_i)$ correspond to the $i$-th (image, label) pair. We construct the task as follows. We randomly select one image to construct the scene-specific unlabeled data $z$. 
Without loss of generality, let us assume that $x_1$ is selected to construct $z$. We feed $z$ to Eq.~\ref{eq:gbn} to compute the GBN parameters $\phi$ from the guiding network parameterized by $\theta$. We then define a loss function that measures the ``goodness'' of the counting network with parameters $\{\phi, \psi\}$. Since our goal is for the learned model to perform well on other images from the same scene, a reasonable loss function is to measure the performance on the remaining $N-1$ images from this scene. We can define the loss for this scene $\mathcal{S}^{t}$ as follows:
\begin{subeqnarray}
  \label{eq:loss_scene}
  \mathcal{L}(\{\psi,\theta\}; \mathcal{S}^t)&&\hspace{-15pt}=\sum_{i=2}^{N}||f(x_i; \{\phi,\psi\})-y_i||^2\\
  &&\hspace{-15pt}=\sum_{i=2}^{N}||f(x_i; \{g(z;\theta), \psi\})-y_i||^2
\end{subeqnarray}
Here we use the $L_2$ distance to measure the difference between the predicted (i.e., $f(x_i; \{g(z;\theta), \psi\})$) and the ground-truth (i.e., $y_i$) density maps. 

The overall objective is to learn $\{\psi,\theta\}$ that minimize Eq.~\ref{eq:loss_scene} across all scenes during training, i.e.,:
\begin{eqnarray}
  \label{eq:loss_all}
  &&\min_{\{\psi,\theta\}}\sum_{t}\mathcal{L}(\{\psi,\theta\}; \mathcal{S}^t)
\end{eqnarray}
In Eq.~\ref{eq:loss_all}, we sum over the loss across all scenes to optimize the parameters during training. 
In Algorithm~\ref{algo:unsupervised_adaptation}, we provide an overview of the learning mechanism.

\begin{algorithm}
	\SetAlgoLined
	\KwIn{Training images from multiple scenes $\{\mathcal{S}^{t}\}$}
	Initialize the model parameters $\{\psi,\theta\}$\;
        \While{not done}{
	  \For{ each scene $\mathcal{S}^{t}$}{
	    Sample one image to obtain $z$\;
            Compute GBN parameters $\phi$ using Eq.~\ref{eq:gbn}\;
            Evaluate $\bigtriangledown_{\{\psi,\theta\}}\mathcal{L}(\{\psi,\theta\};\mathcal{S}^t)$ in Eq.~\ref{eq:loss_scene} \;
	  }
          Update $\{\psi,\theta\}\leftarrow \sum_{t}\bigtriangledown_{\{\psi,\theta\}} \mathcal{L}(\{\psi,\theta\};\mathcal{S}^t)$ \;
	}
	\caption{Training for unlabeled scene-adaptive crowd counting}
	\label{algo:unsupervised_adaptation}
\end{algorithm}

Let $\{\psi^*, \theta^*\}$ be the learned model parameters that minimize Eq.~\ref{eq:loss_all}. During testing, we are given a new scene represented as $z_{new}$ (unlabeled image). For any crowd image $x$ from this scene, we can predict its label $\hat{y}$ as:
\begin{equation}
  \hat{y}=f(x; \{\phi^*, \psi^*\}), \textrm{ where } \phi^*=g(z_{new};\theta^*)
\end{equation}

The idea of using adjustable GBN parameters for adaptation is inspired by the work in image translation~\cite{liu2019few}. Similar to \cite{liu2019few}, the affine transformation in the normalization layers is spatially invariant, so it can only obtain global appearance information. In the context of crowd counting, different scenes are often characterized by some factors (e.g., camera angle, height) that cause changes in the global scene appearance. By adapting parameters in the GBN layers (which are spatially invariant), the parameter $\phi$ obtained via the guiding network from the unlabeled images intuitively captures the global information (e.g., scene geometry) about the target scene. The local structures needed in crowd counting are implicitly captured by the parameter $\psi$ and are shared across different scenes. 
\section{Experiments}
In this section, we first describe the datasets and the experiment setup in Sec.~\ref{sec:dataset}. We then introduce several baseline methods used for comparisons in Sec.~\ref{sec:baseline}. We present experimental results in Sec.~\ref{sec:results}.

\subsection{Datasets and Setup}\label{sec:dataset}
\noindent\textbf{Datasets:} We experiment with the following datasets.
\begin{itemize}
	\item \textit{WorldExpo'10}~\cite{zhang15cross}: The WorldExpo'10 dataset consists of 3980 labeled images covering 108 different surveillance camera scenes. The dataset is split into training set (3380 images over 103 scenes) and testing set (600 images over 5 scenes).  Note that although the dataset comes with ROI maps, for the sake of simplicity,  we do not use them in our experiments as the primary objective of our work is to show the benefit of unlabeled scene adaptation.  In our experiments, we resize the images to 512 $\times$ 672. 
	\item \textit{Mall}~\cite{change2013semicounting}: The Mall dataset consists of 2000 frames captured from a surveillance camera in a mall. The frame size is 640 $\times$ 480. The training and test sets consist of 800 and 1200 images, respectively.
	\item \textit{PETS}~\cite{ferryman2009pets2009}: The PETS dataset is a multi-view dataset of crowd scene from 8 views. We use the first 3 views following~\cite{zhang2019wide} and similarly we consider sequences S1L3 (14\_17, 14\_33), S2L2 (14\_55) and S2L3 (14\_41) as the training set consisting of 1105 images. For the test set, we consider S1L1 (13\_57, 13\_59), S1L2 (14\_06, 14\_31) consisting of 794 images. We treat each view as a scene in our experiments and hence we consider 3 scenes in~\cite{ferryman2009pets2009}. The original image size is 576 $\times$ 768. In our experiments, we resize the images to 288 $\times$ 384 following~\cite{zhang2019wide}.
    \item \textit{FDST}~\cite{fang2019locality}: The FDST dataset is made up of 13 different camera scenes. The training set consists of 60 videos resulting in 9000 frames and the testing set consists of 40 videos resulting in 6000 frames. The dataset contains images of two different resolutions (1920 $\times$ 1080 or 1280 $\times$ 720). Following~\cite{fang2019locality}, we resize all the frames to 640 $\times$ 360 in our experiments.
    \item \textit{CityUHK-X}~\cite{kang2017incorporating}: The CityUHK-X dataset consists of a total of 55 camera scenes. The training set comprises of 43 scenes and the testing set comprises of the remaining 12 scenes. The frame size is 384 $\times$ 512.
    \item \textit{Venice}~\cite{liu2019context}: The Venice dataset comprises of 4 different crowd sequences captured from a mobile camera. The dataset contains 167 labeled frames in total. The training set consists of 80 images from one single sequence. The test set consists of frames from the other three video sequences. We treat frames of each test video as a different scene for scene adaptation.
\end{itemize}

Our problem setup requires data from multiple scenes during the training phase. To the best of our knowledge, WorldExpo'10~\cite{zhang15cross} is the only real-world crowd counting dataset with a large number of scenes that can be used for training in our problem setup. Therefore, we use WorldExpo'10 for training and use other datasets to test for scene adaptation. \\

\begin{table*}[ht]
	\centering
        \caption{Quantitative results for training and testing on WorldExpo'10. ``Ours'' uses one unlabeled image $z$. We report results using different backbone architectures, including CSRNet~\cite{li2018csrnet}, FCN~\cite{he2016deep}, and SFCN~\cite{wang2019learning}. The results for our approach show mean and standard deviation (\%) over 5 random trials. We show the best results in \textbf{bold}.}
	\renewcommand{\arraystretch}{1.1}
	\begin{tabular}{llc|cc}
		\hline
		Backbone & Method	 & Adaptive & MAE ($\downarrow$)   & RMSE ($\downarrow$) \\ \hline
		 & CSRNet                & - &19.56     &28.34  \\ 
		 VGG-16 & CSRNet w/ BN       & - & 18.57    & 29.91 \\ 
		 & \textbf{Ours w/ CSRNet}   & \checkmark & \textbf{17.32}~\smaller{$\pm{0.1}$}     & \textbf{27.03} \smaller{$\pm {0.05}$}  \\ \hline
		 
		 & FCN              & - & 23.07    & 34.16  \\ 
		 ResNet-101 & FCN w/ BN          & - & 21.65    & 33.3  \\ 
		 & \textbf{Ours w/ FCN}  & \checkmark & \textbf{20.91} \smaller{$\pm {0.3}$}    & \textbf{29.61} \smaller{$\pm {0.2}$}   \\ \hline
		 
		 & SFCN         & - &   25.47  & 35.91    \\ 
		ResNet-101 & SFCN w/ BN   & - & 23.84   & 35.52 \\  
		 & \textbf{Ours w/ SFCN} & \checkmark & \textbf{14.56} \smaller{$\pm {0.4}$}    & \textbf{22.75} \smaller{$\pm {0.2}$}  \\ 
			\hline
	\end{tabular}
    \label{tab:we}
\end{table*}

\begin{table*}
	\centering
	\caption{Quantitative results for the cross-dataset testing for one unlabeled image. We train on WorldExpo'10 and test on Mall, PETS, FDST, and CityUHK-X. We show results of different backbone networks and report mean and standard deviation (\%) of our models over 5 random trials. We show the best results in \textbf{bold}.}
	\smaller
	\renewcommand{\arraystretch}{1.2}
	\begin{tabular}{lc|cc|cc|cc|cc}
		\hline
		\multirow{2}{*}{Method} & \multirow{2}{*}{Adapt.} &\multicolumn{2}{c}{WorldExpo $\rightarrow$ Mall} & \multicolumn{2}{c}{WorldExpo $\rightarrow$ PETS} & \multicolumn{2}{c}{WorldExpo $\rightarrow$ FDST} & \multicolumn{2}{c}{WorldExpo $\rightarrow$ CityUHK-X}\\ 
	\cline{3-10}
		& & MAE ($\downarrow$) & RMSE ($\downarrow$) & MAE ($\downarrow$) & RMSE ($\downarrow$) & MAE ($\downarrow$) & RMSE ($\downarrow$) & MAE ($\downarrow$) & RMSE ($\downarrow$)
                \\ \hline
		CSRNet ~\cite{li2018csrnet}  & - & 9.94    & 10.41    &  17.99   &  19.80  &       12.74    &  13.09 & 39.85   & 48.07 \\ 
		CSRNet w/ BN~\cite{li2018csrnet}  & - & 8.72    & 9.92      &  18.63   &  20.49     &   7.30 &  7.81 & 22.69	& 32.33 \\ 
		\textbf{Ours w/ CSRNet}  & \checkmark & \textbf{4.0} \smaller{$\pm {0.08}$}    & \textbf{5.0} \smaller{$\pm {0.09}$}    &  \textbf{17.43} \smaller{$\pm {0.04}$}   &  \textbf{19.70} \smaller{$\pm {0.05} $}       &  \textbf{7.14} \smaller{$\pm {0.3} $}   &   \textbf{7.77} \smaller{$\pm {0.2} $}   &  \textbf{20.38} \smaller{$\pm {0.004} $}   &   \textbf{29.06} \smaller{$\pm {0.003} $} \\ \hline
		
		FCN~\cite{he2016deep}   &  - & 9.03    & 9.6     &  20.38   & 22.67  &  7.14   &  7.85  & 27.56	& 37.73\\ 
		FCN w/ BN~\cite{he2016deep} & - & 9.63    & 10.32   &  19.59   &  21.61     &  8.89   &  9.23        & 27.3	& 37.56\\ 
		\textbf{Ours w/ FCN}  & \checkmark &  \textbf{4.12} \smaller{$\pm {0.2}$}   & \textbf{5.12} \smaller{$\pm {0.2}$}  & \textbf{13.74} \smaller{$\pm {0.1}$}     & \textbf{16.15} \smaller{$\pm {0.09}$}     &  \textbf{6.13} \smaller{$\pm {0.4} $}   & \textbf{6.69} \smaller{$\pm {0.3} $}  & \textbf{21.77} \smaller{$\pm {0.012} $}   & \textbf{30.94} \smaller{$\pm {0.0124} $}\\ \hline
		
		SFCN~\cite{wang2019learning}     & - &  15.17   &  15.53  &  19.62   &  21.55     &  8.72 & 8.94  & 27.34 & 35.78 \\ 
		SFCN w/ BN~\cite{wang2019learning}  & - & 10.8    &  11.39 &  19.94   &  22.26  &  6.60   &  6.96  & 26.15 & 36.61\\  
		\textbf{Ours w/ SFCN}  & \checkmark & \textbf{6.99} \smaller{$\pm {1.2}$}   & \textbf{8.0} \smaller{$\pm {1.0}$}  & \textbf{18.41} \smaller{$\pm {0.2} $}  & \textbf{20.63} \smaller{$\pm {0.1} $} &  \textbf{5.76} \smaller{$\pm {0.3} $}  &  \textbf{6.57 $\pm {\tiny 0.3} $} &  \textbf{20.14} \smaller{$\pm {0.006} $}  &  \textbf{28.27 $\pm {\tiny 0.013} $} \\ 
		\hline
	\end{tabular} 
	\label{tab:we_other}
\end{table*}

\noindent\textbf{Implementation Details:} We implement the {\fontfamily{qcr}\selectfont AdaCrowd} framework with different backbone (VGG-16~\cite{simonyan2014very} or ResNet-101~\cite{he2016deep}) networks.  The crowd counting backbone encoders are initialized with ImageNet~\cite{deng2009imagenet} pre-trained weights,  while the weights for the guiding network and crowd counting decoder are randomly initialized.  The learning rate for all the experiments is set to $\texttt{1e-5}$. We use Adam~\cite{kingma2015adam} optimizer with gradient clipping norm of 1. We train all models for 110 epochs with a batch size of 1 (i.e., number of images to the crowd counting encoder) and vary the unlabeled images to the guiding network (e.g., 1 and 5).  We use $\texttt{StepLR}$ learning rate scheduler with a $\texttt{step\_size}$ of 1 and $\texttt{gamma}$ of 0.995.  We have implemented all our networks in PyTorch~\cite{paszke2017automatic}.  For details on the network architectures,  please refer to the supplementary material.  \\ 

\noindent\textbf{Evaluation Metrics:} Following previous work~\cite{li2018csrnet, zhang15cross, Reddy_2020_WACV}, we use two standard evaluation metrics, namely mean absolute error (MAE) and root mean square error (RMSE), as shown below:
\begin{subeqnarray}
&&MAE = \dfrac{1}{N}\sum_{i=1}^{N}|C(\hat{y}_i) - C(y_i)|\\
&&RMSE = \sqrt{\dfrac{1}{N}\sum_{i=1}^{N}|C(\hat{y}_i) - C(y_i)|^{2}}
\end{subeqnarray}
where $N$ represents the number of test images. We denote the ground-truth and the predicted density map for a given $i$-th input image as ${y}_{i}$ and $\hat{y}_{i}$, respectively. We use $C(\cdot)$ to denote the total count by summing over all values of a density map. Given a density map $y$, we use $y[h, w]$ to denote the value at a particular spatial location $[h, w]$ in the density map $y$. The total crowd count in a density map with spatial resolution $H \times W$ can be computed by $C(y) = \sum_{h=1}^{H}\sum_{w=1}^{W}{y[h,w]}$. For MAE and RMSE, lower values indicate better performance.

\subsection{Baselines and Backbone Architectures}\label{sec:baseline}
We consider the following state-of-the-art crowd counting networks as baselines for comparison. These baselines train the networks in the standard supervised manner on images available during training, then directly apply the trained networks to test different scenes without adaptation. 

\begin{itemize}
	\item \textit{CSRNet}~\cite{li2018csrnet}: The first sub-network in CSRNet is based on VGG~\cite{simonyan2014very} and consists of layers till $\texttt{conv\_4\_3}$ to extract the features with dimension $H/8 \times W/8 \times 512$ from the image input. The extracted features are used to construct the density map using a series of dilated convolutional layers in the second sub-network.  Note that the original CSRNet paper uses ROI maps to improve the performance. We use the CSRNet without the ROI maps for simplicity, since the ROI maps are not already available in practice.

	\item {\textit{CSRNet w/ Batch Normalization}~\cite{li2018csrnet}: } This baseline is based on CSRNet. We add batch normalization layers between every $conv$ and $relu$ layer in the second sub-network to generate the density map.

	\item {\textit{ResNet FCN}~\cite{wang2019learning}: } This baseline is based on the ResNet-101~\cite{he2016deep} architecture to extract the image features with dimension $H/8 \times W/8 \times 1024$. Similar to CSRNet, we first extract the image features and then generate the density map using the second sub-network comprising of dilated convolutional layers.

	\item {\textit{ResNet FCN w/ Batch Normalization}~\cite{wang2019learning}:} This baseline is similar to ResNet FCN. We add batch normalization layers to the density map generator sub-network like in CSRNet w/ BN.

	\item {\textit{ResNet SFCN}~\cite{wang2019learning}:} This baseline uses ResNet-101~\cite{he2016deep} for the feature extractor part of the network. The density map generator consists of dilated convolutional layers and spatial fully connected layers as proposed in~\cite{wang2019learning}.

	\item {\textit{ResNet SFCN w/ Batch Normalization}~\cite{wang2019learning}:} This baseline is based on ResNet SFCN. We add batch normalization layers in the density map generator like in other BN based baselines.
	
\end{itemize}

We use \textit{CSRNet}, \textit{ResNet FCN} and \textit{ResNet SFCN} (with GBN layers inserted) as the backbone architecture in our {\fontfamily{qcr}\selectfont AdaCrowd} framework. To be specific, we can derive an {\fontfamily{qcr}\selectfont AdaCrowd} variant for each of the BN-based baseline methods by replacing all BN layers with GBN layers. In our approach, we generate the parameters for the GBN layers using the guiding network based on the unlabeled data $z$. 

\begin{figure*}[!t]
  \centering
  {\setlength{\tabcolsep}{0.5pt}
        \begin{tabular}{ccccc}
        
CSRNet~\cite{li2018csrnet} &
				\raisebox{-.5\height}{
          		\includegraphics[height=1.3in, width=1.4in]{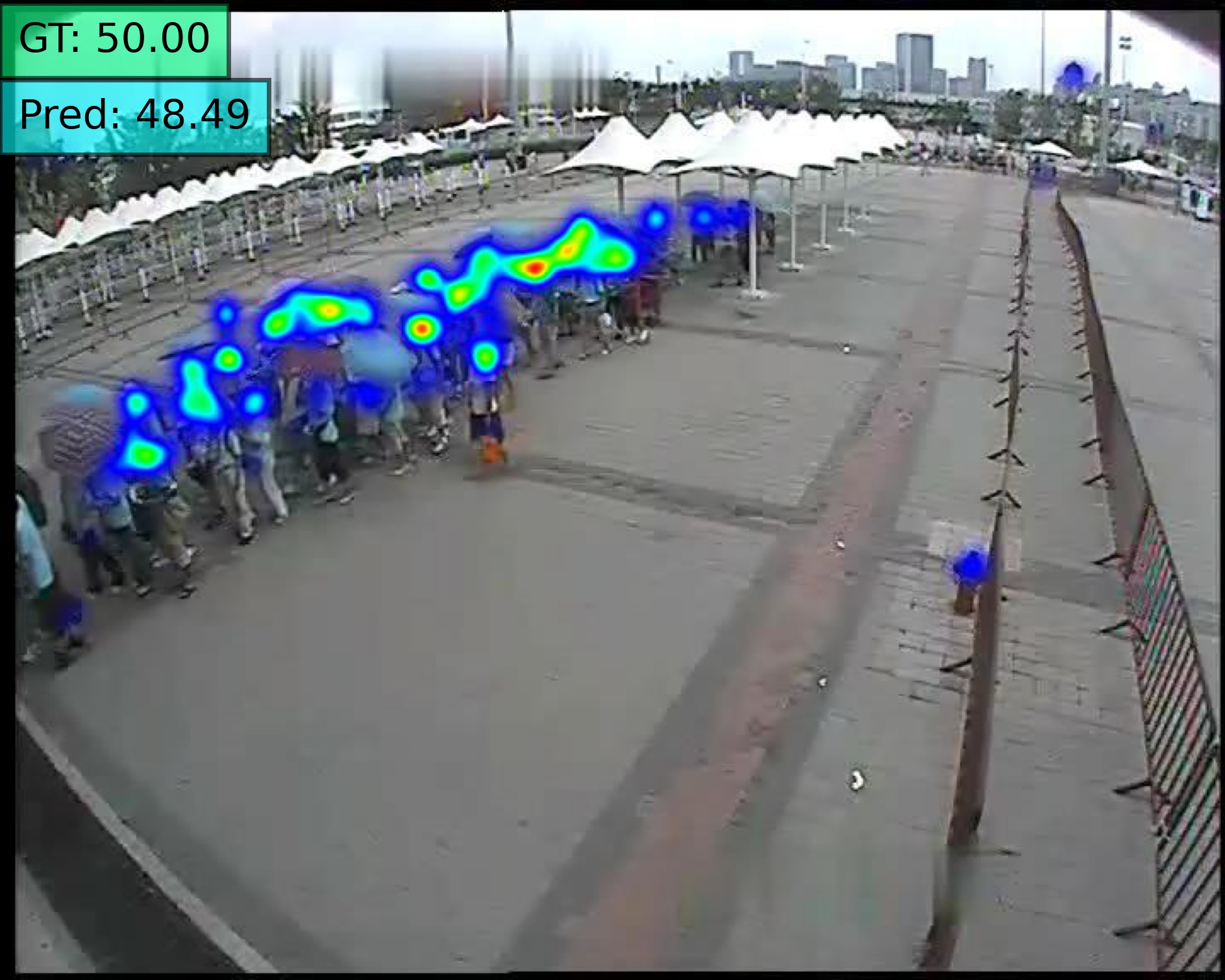}}&
          		\raisebox{-.5\height}{
          	\includegraphics[height=1.3in, width=1.4in]{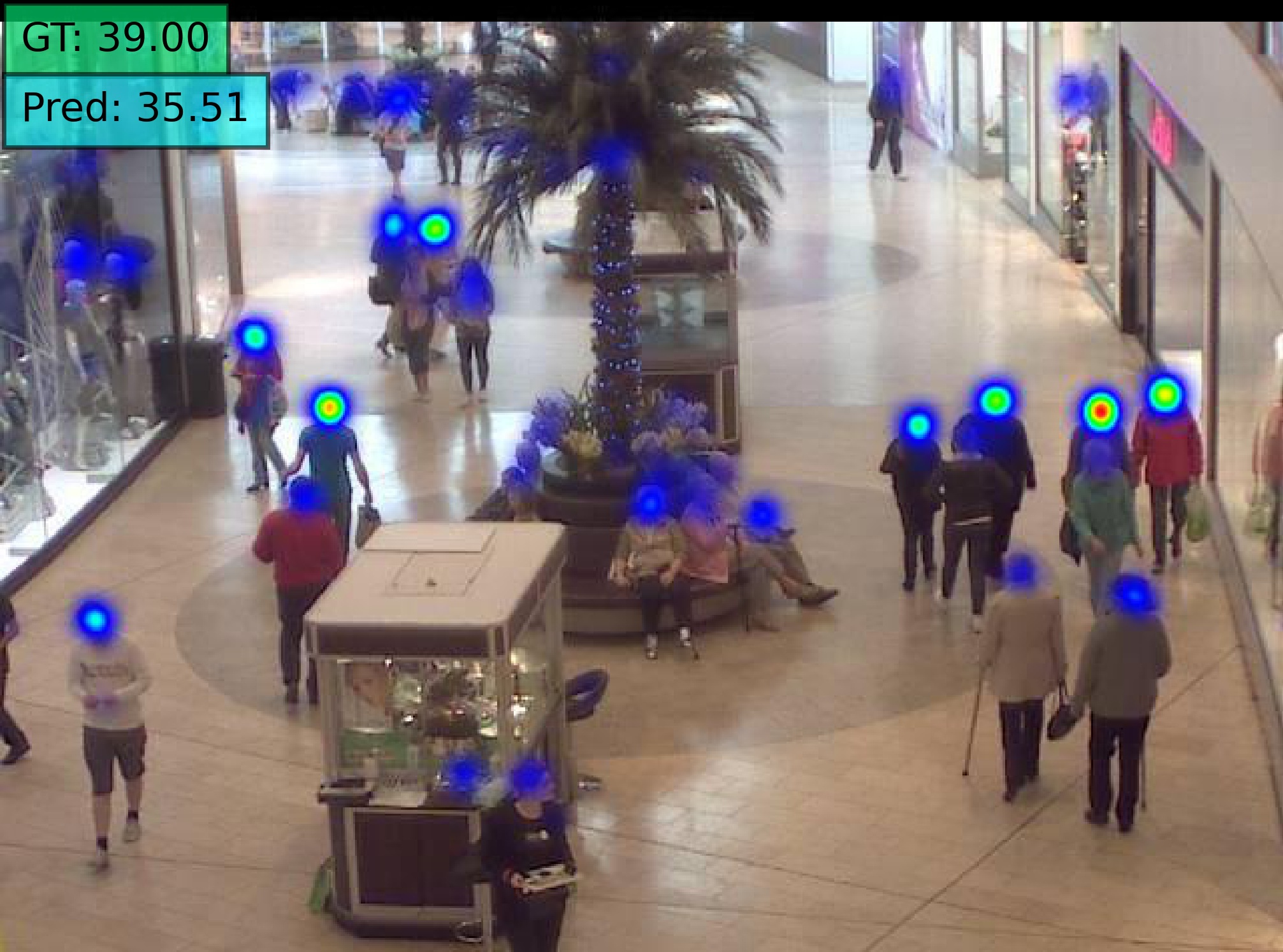}}&
          	\raisebox{-.5\height}{
          	\includegraphics[height=1.3in, width=1.4in]{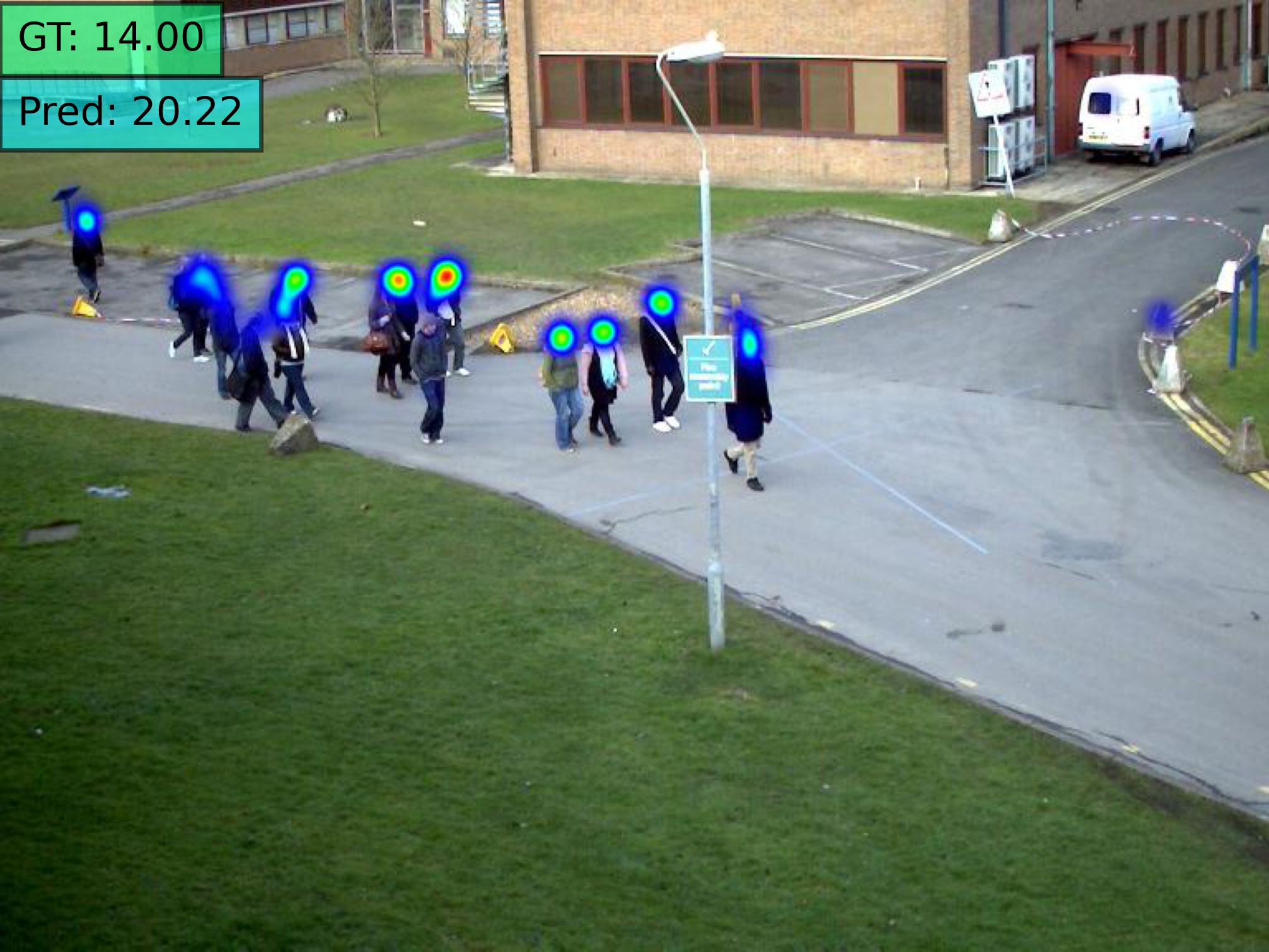}}&
          	\raisebox{-.5\height}{
          	\includegraphics[height=1.3in, width=1.4in]{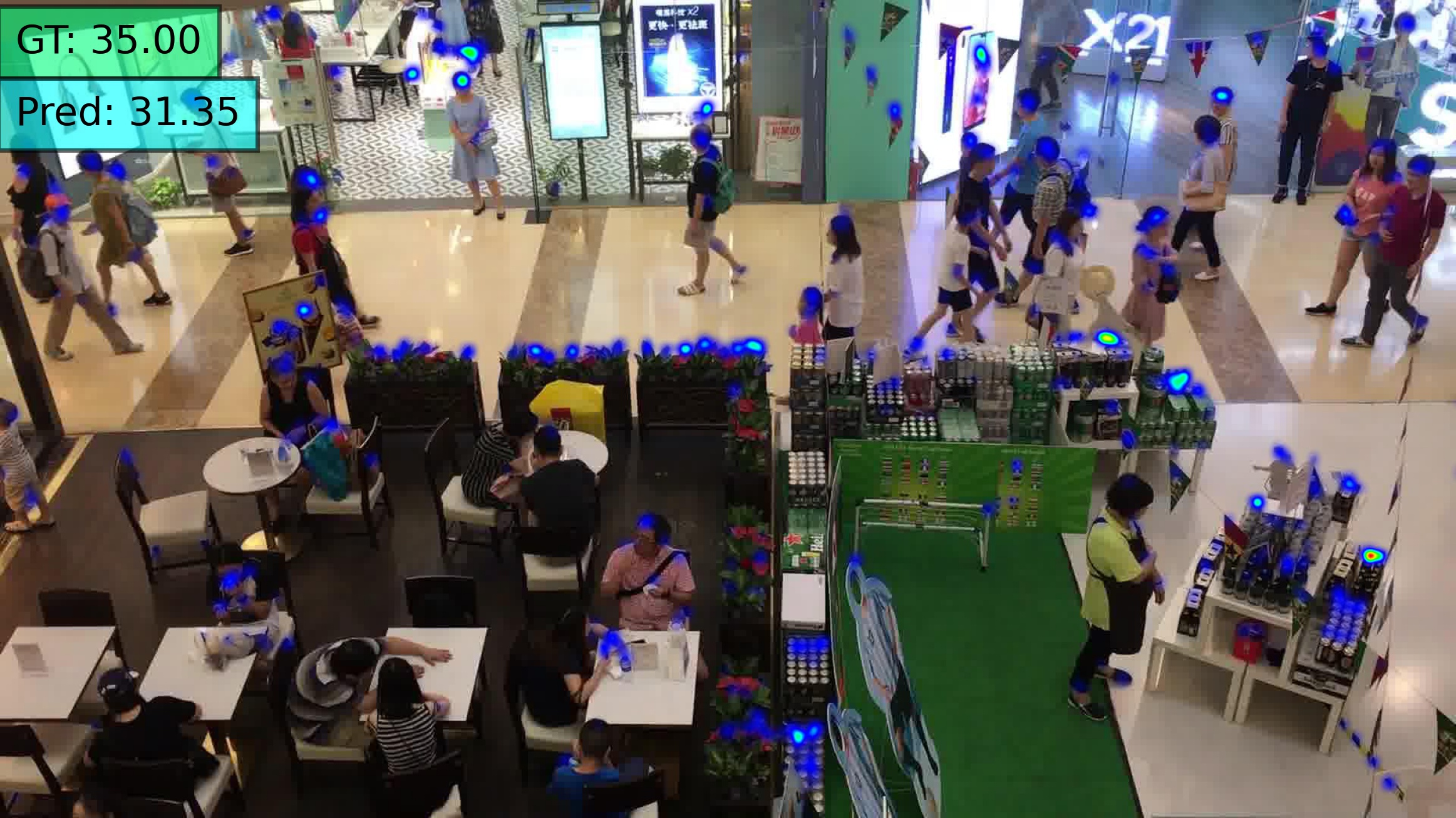}}\\
          	
CSRNet w/ BN~\cite{li2018csrnet} &
		\raisebox{-.5\height}{
        	\includegraphics[height=1.3in, width=1.4in]{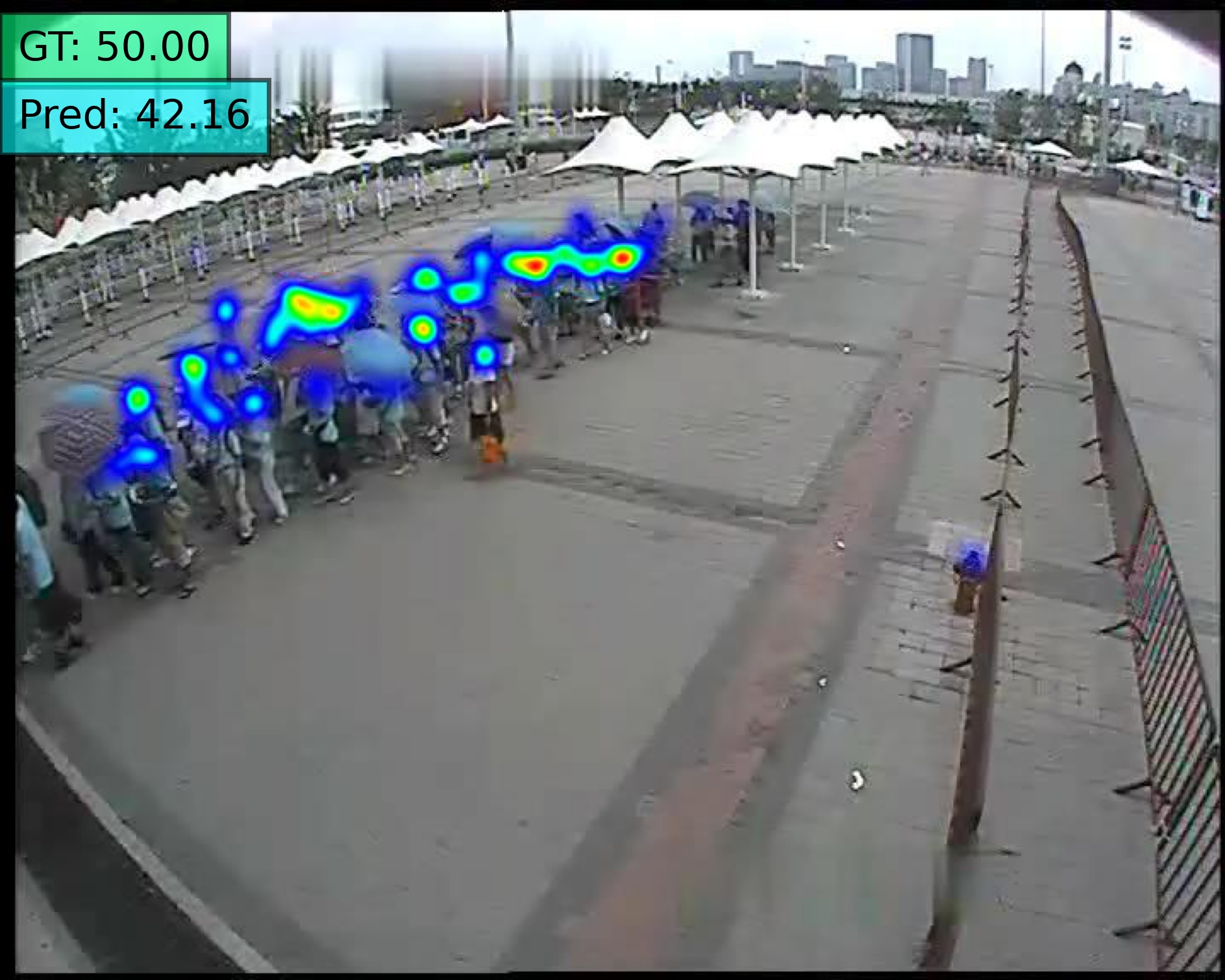}}&
        	\raisebox{-.5\height}{
          	\includegraphics[height=1.3in, width=1.4in]{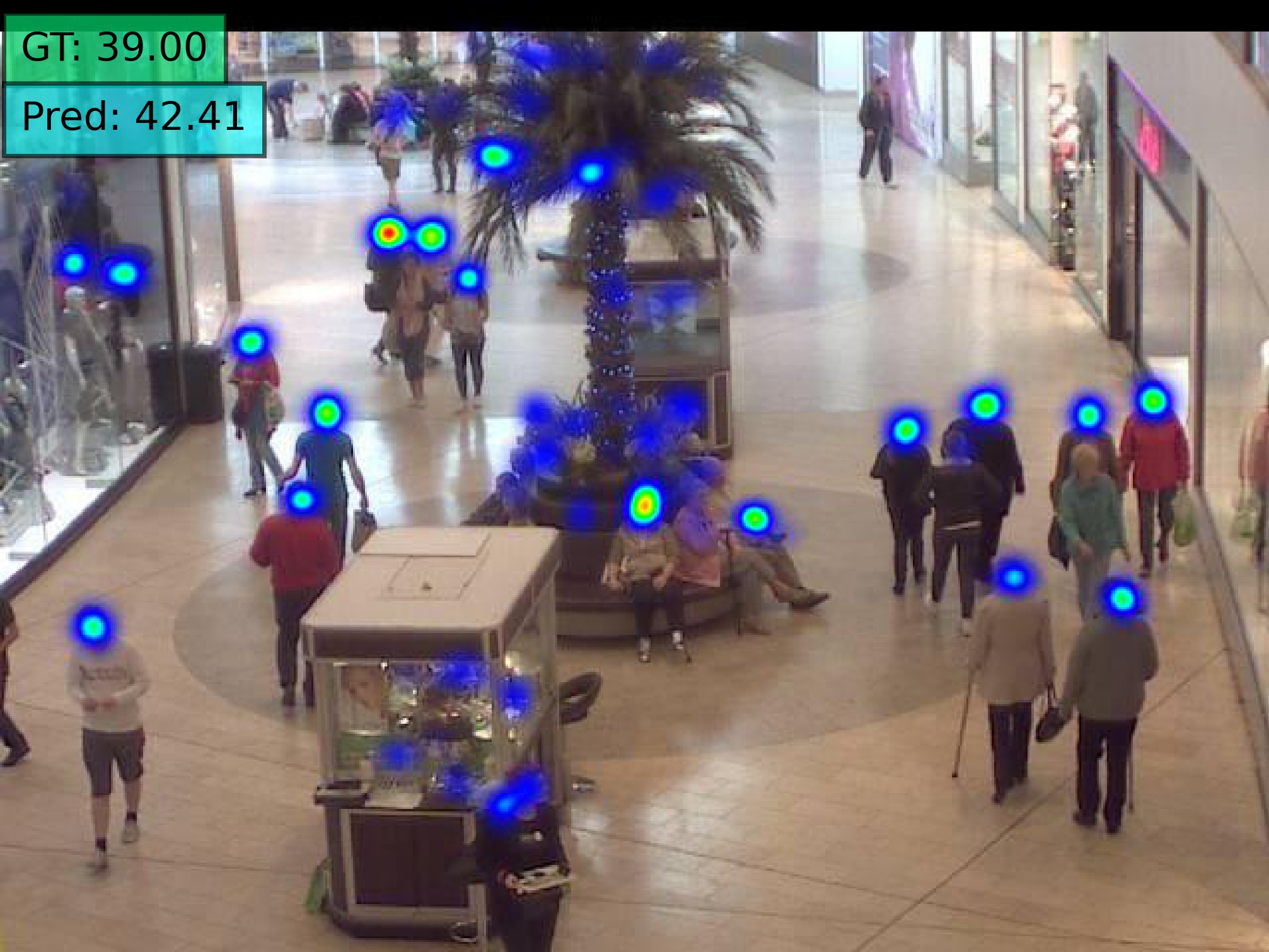}}&
          	\raisebox{-.5\height}{
          	\includegraphics[height=1.3in, width=1.4in]{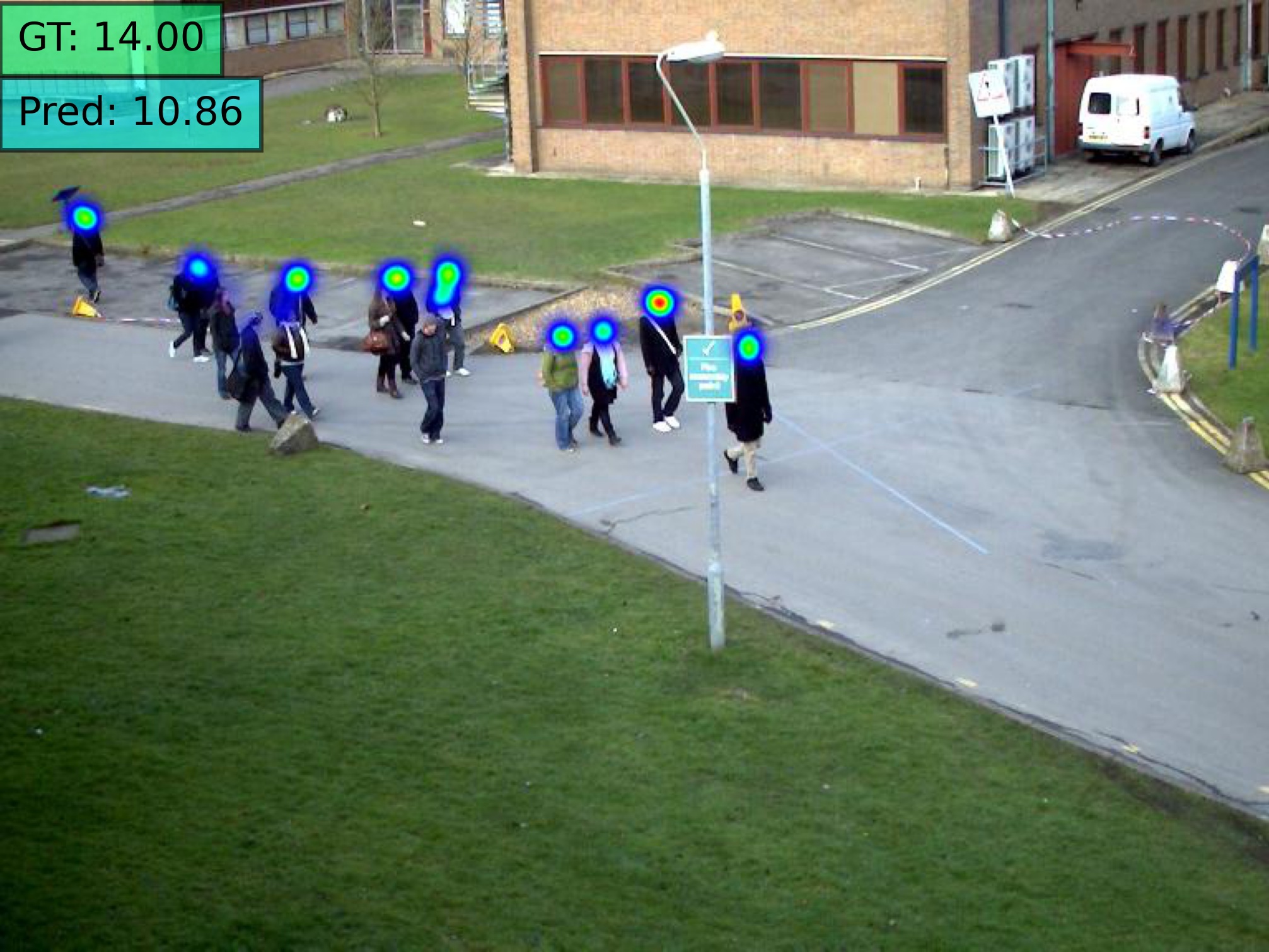}}&
          	\raisebox{-.5\height}{
          	\includegraphics[height=1.3in, width=1.4in]{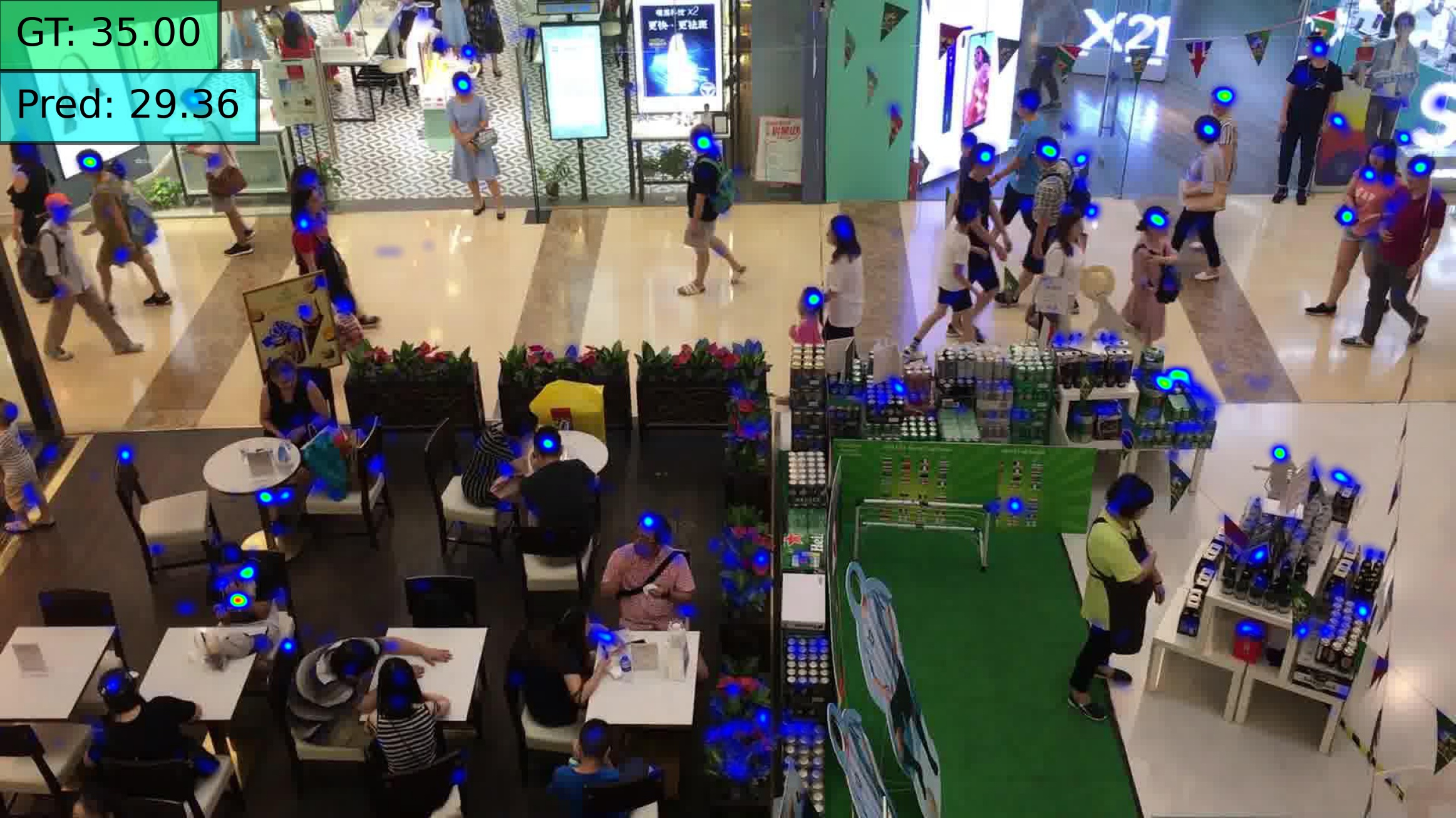}}\\
          	
		\textbf{Ours w/ CSRNet} &
		\raisebox{-.5\height}{
        	\includegraphics[height=1.3in, width=1.4in]{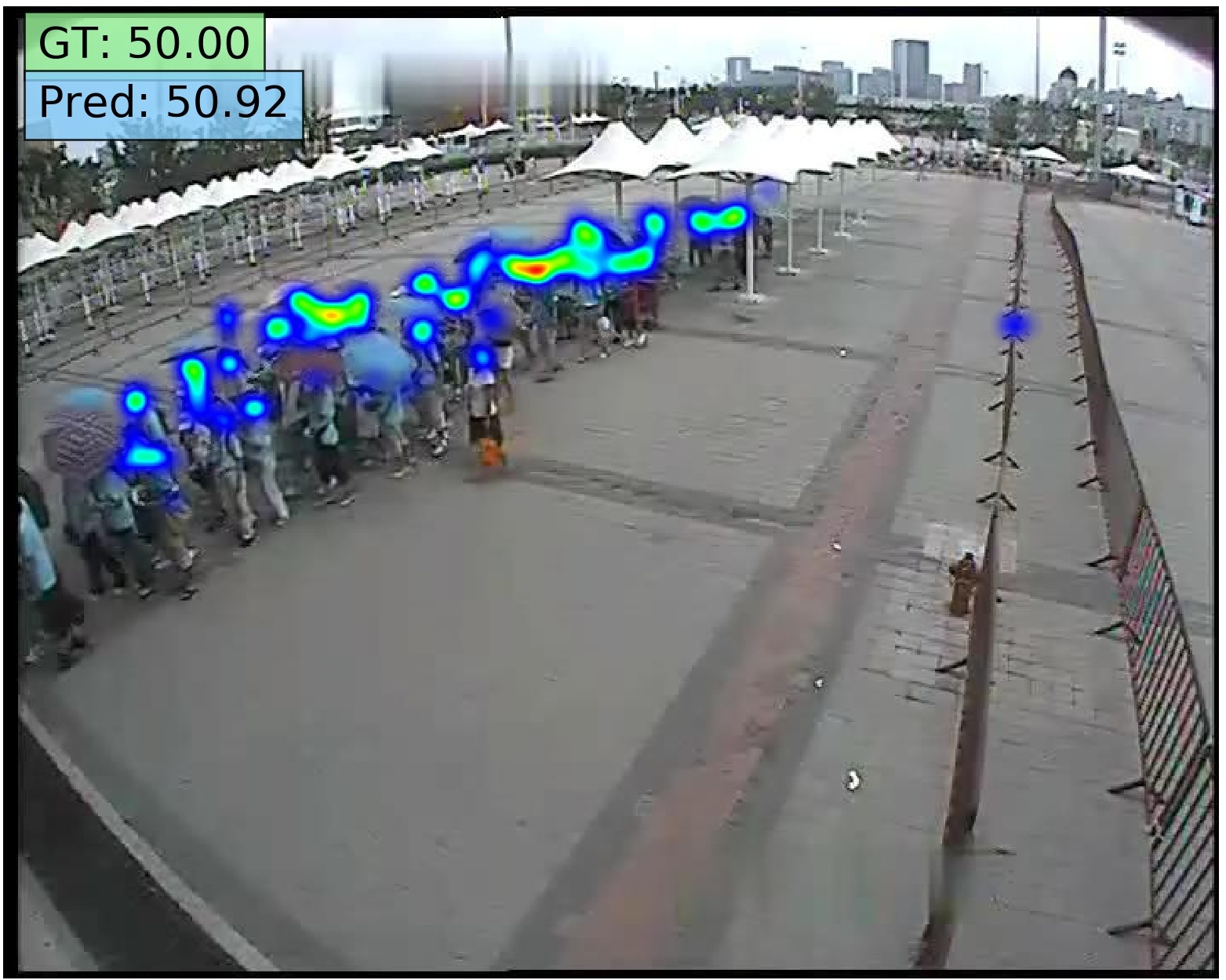}}&
        	\raisebox{-.5\height}{
          	\includegraphics[height=1.3in, width=1.4in]{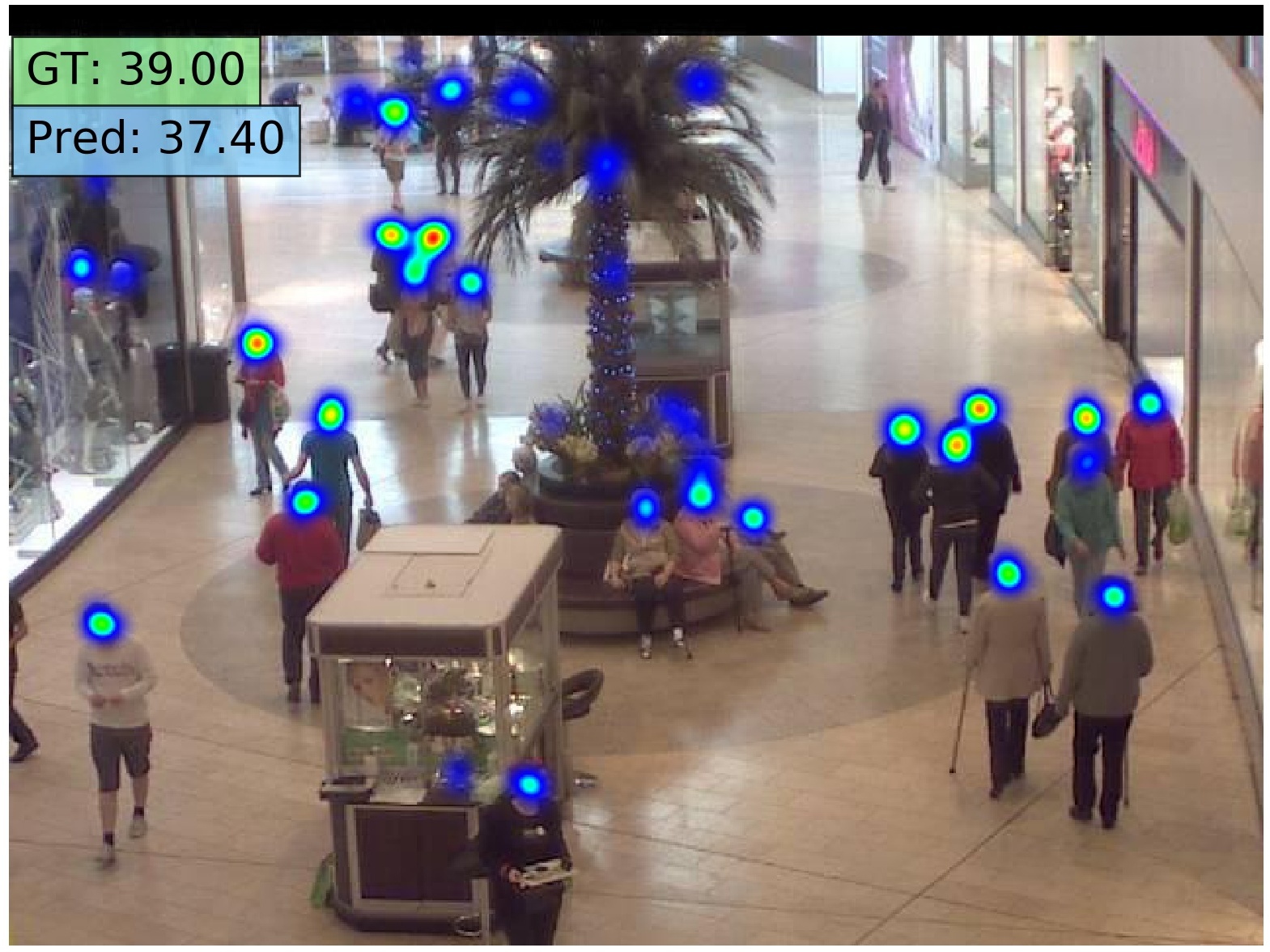}}&
          	\raisebox{-.5\height}{
          	\includegraphics[height=1.3in, width=1.4in]{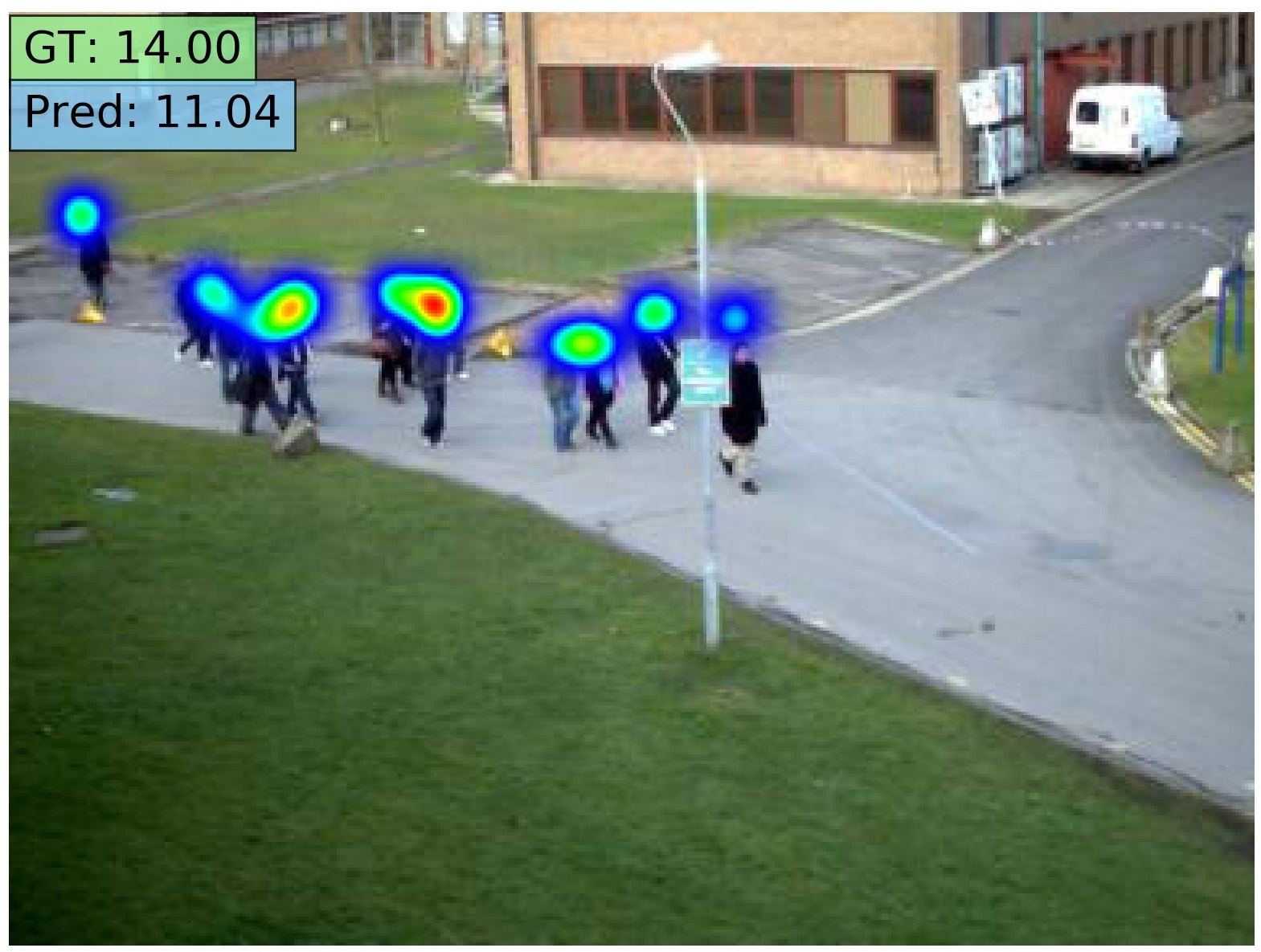}}&
          	\raisebox{-.5\height}{
          	\includegraphics[height=1.3in, width=1.4in]{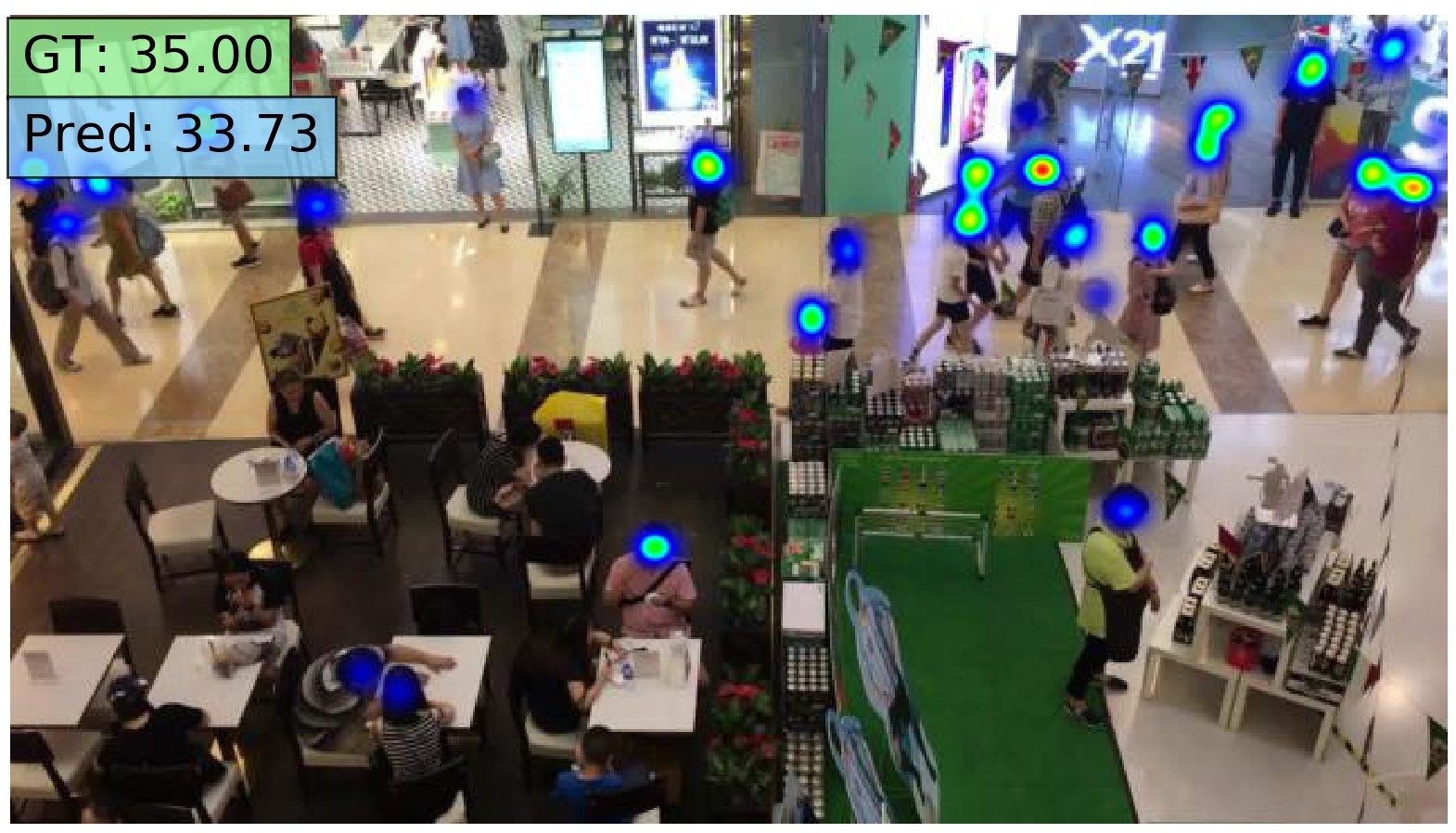}}\\
          	

CSRNet~\cite{li2018csrnet} &
			\raisebox{-.5\height}{
         	\includegraphics[height=1.3in, width=1.4in]{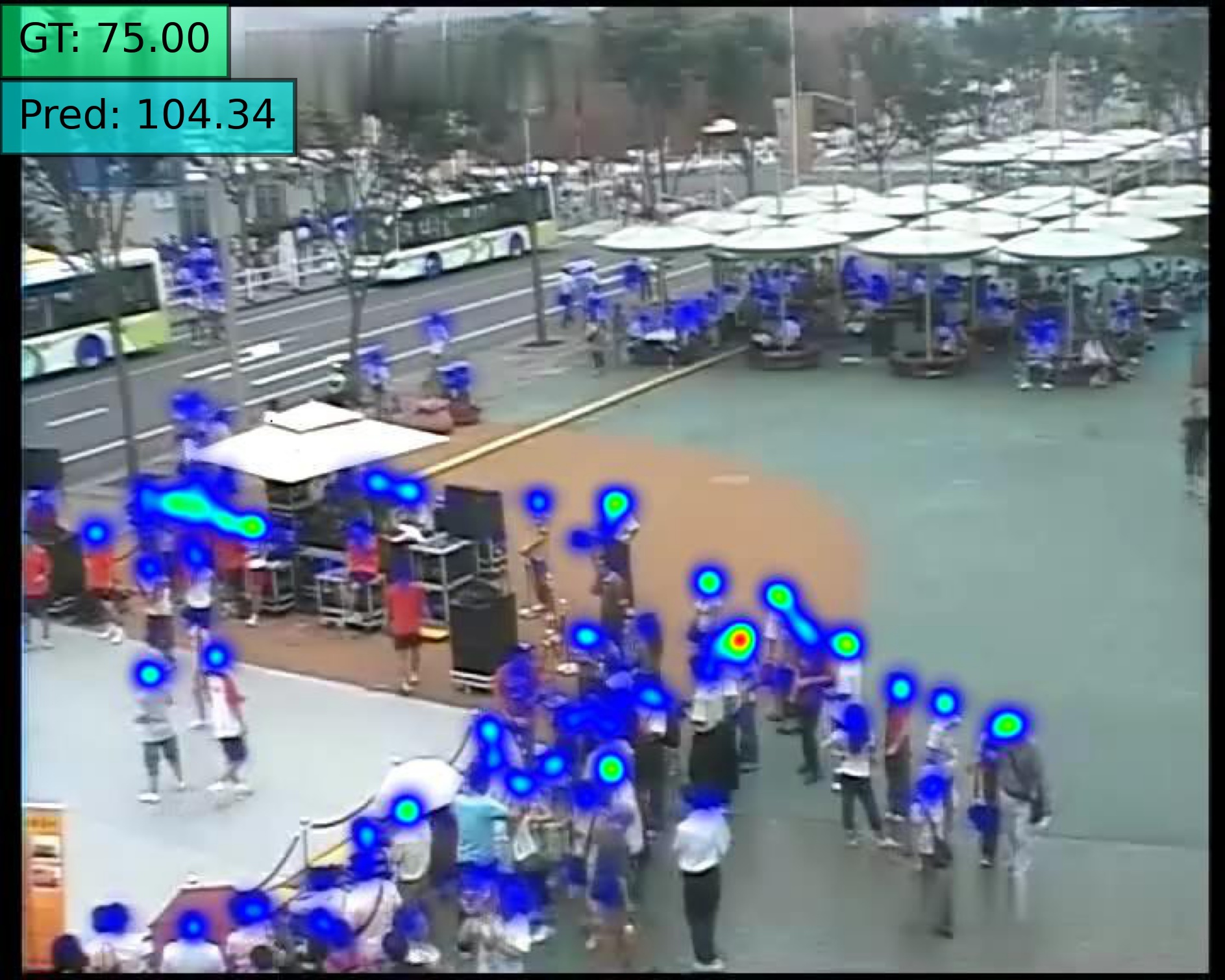}}&
         	\raisebox{-.5\height}{
          	\includegraphics[height=1.3in, width=1.4in]{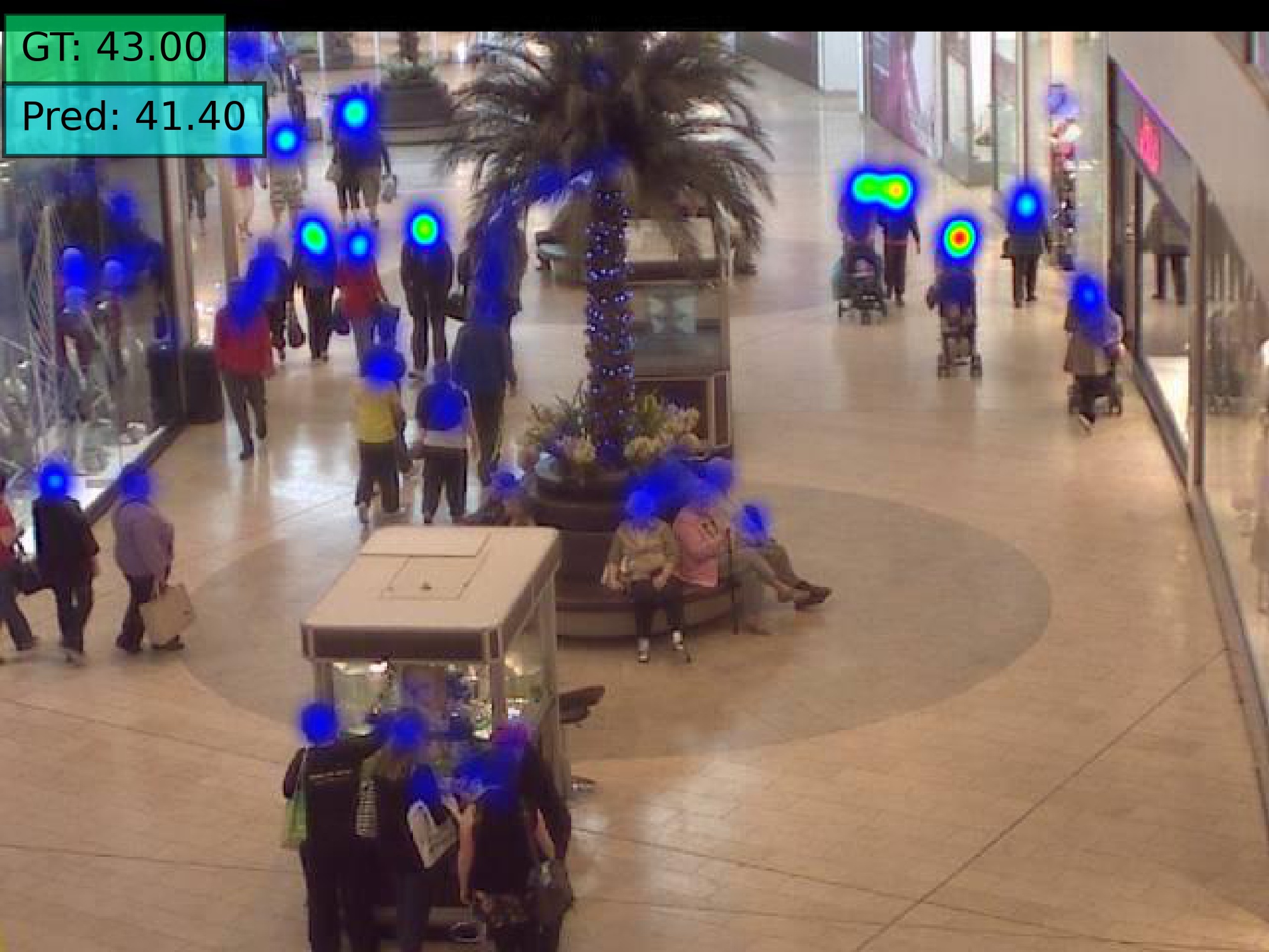}}&
          	\raisebox{-.5\height}{
          	\includegraphics[height=1.3in, width=1.4in]{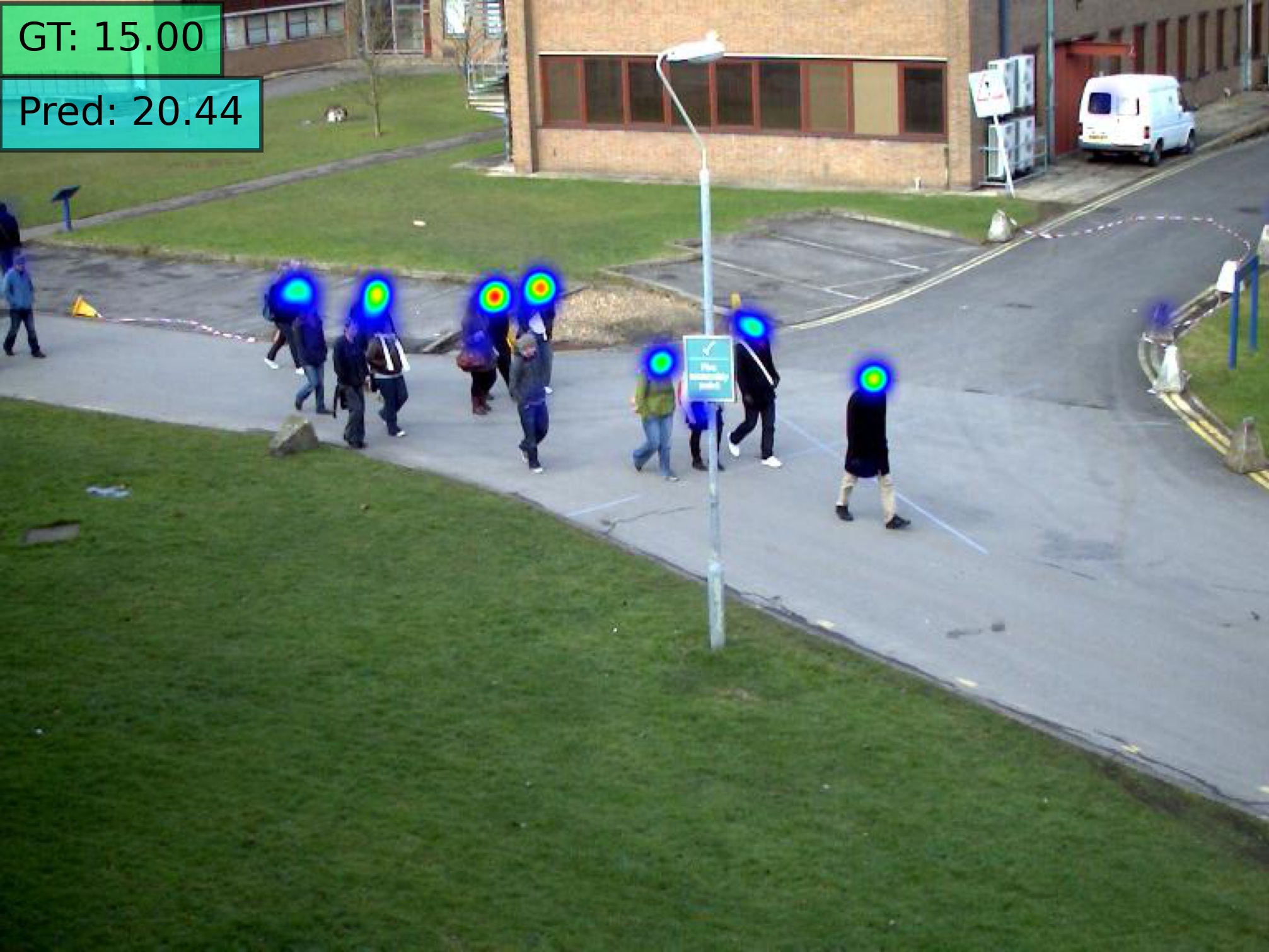}}&
          	\raisebox{-.5\height}{
          	\includegraphics[height=1.3in, width=1.4in]{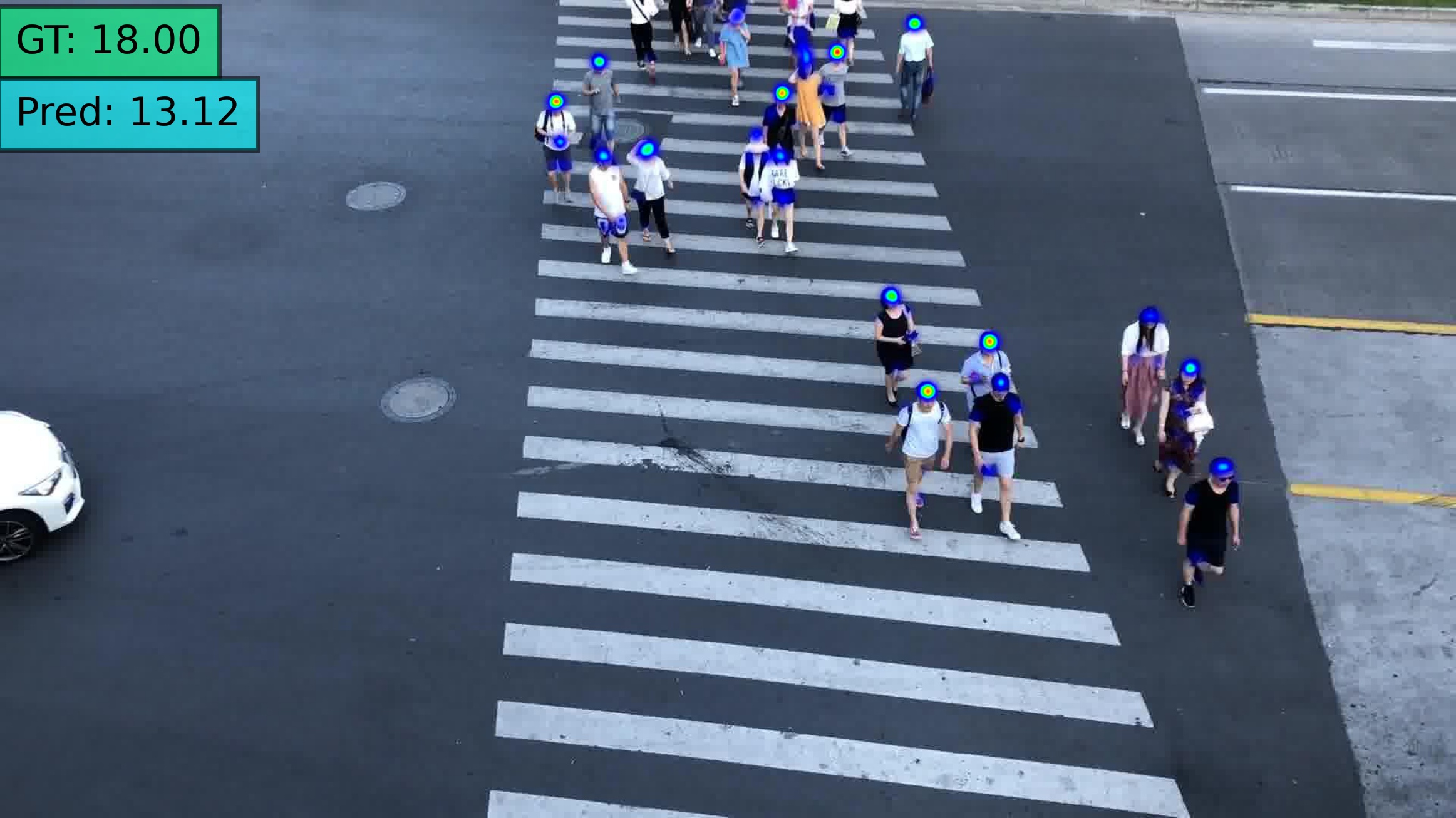}}\\
          	
CSRNet w/ BN~\cite{li2018csrnet} &
			\raisebox{-.5\height}{         
          	\includegraphics[height=1.3in, width=1.4in]{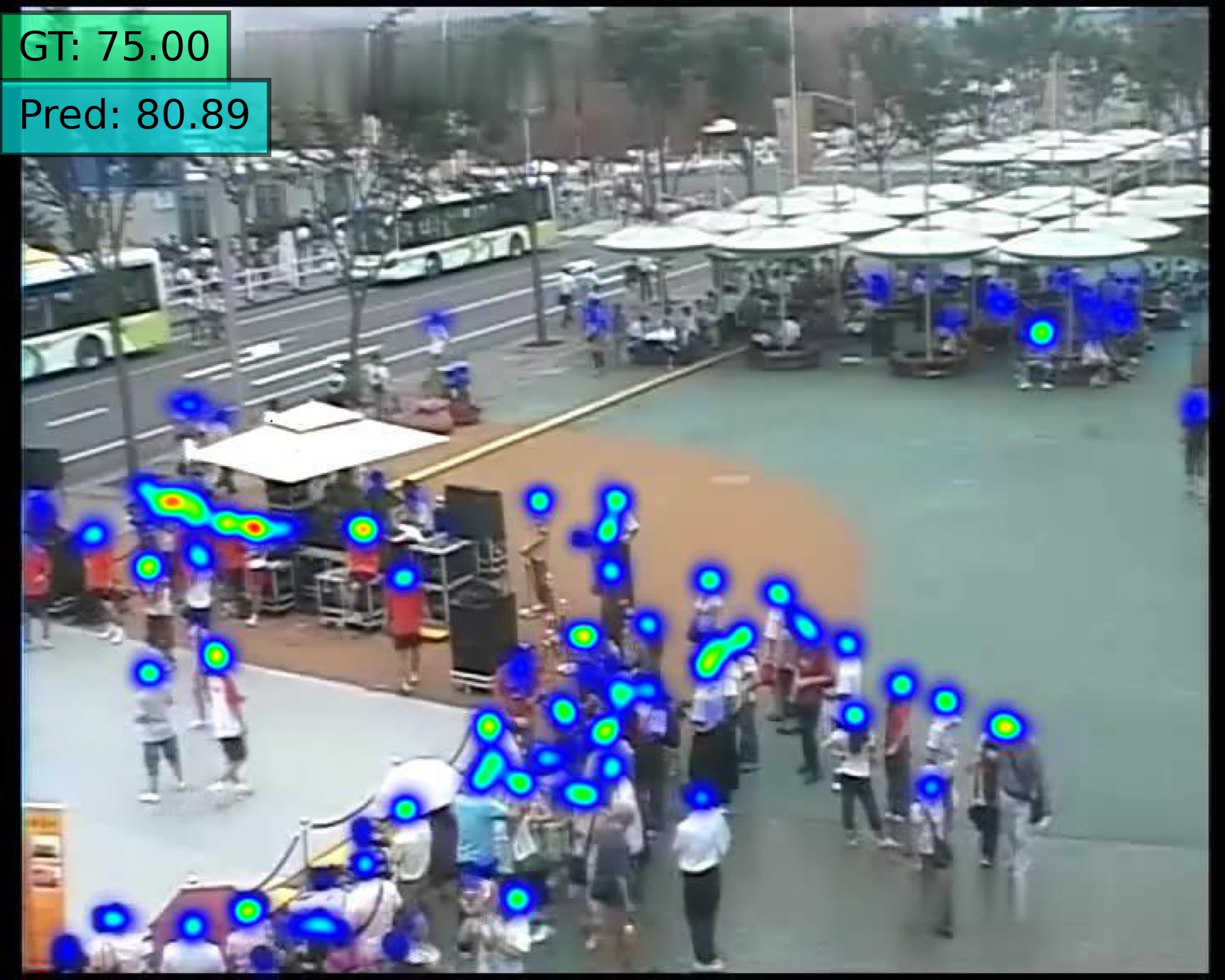}}&
          	\raisebox{-.5\height}{
          	\includegraphics[height=1.3in, width=1.4in]{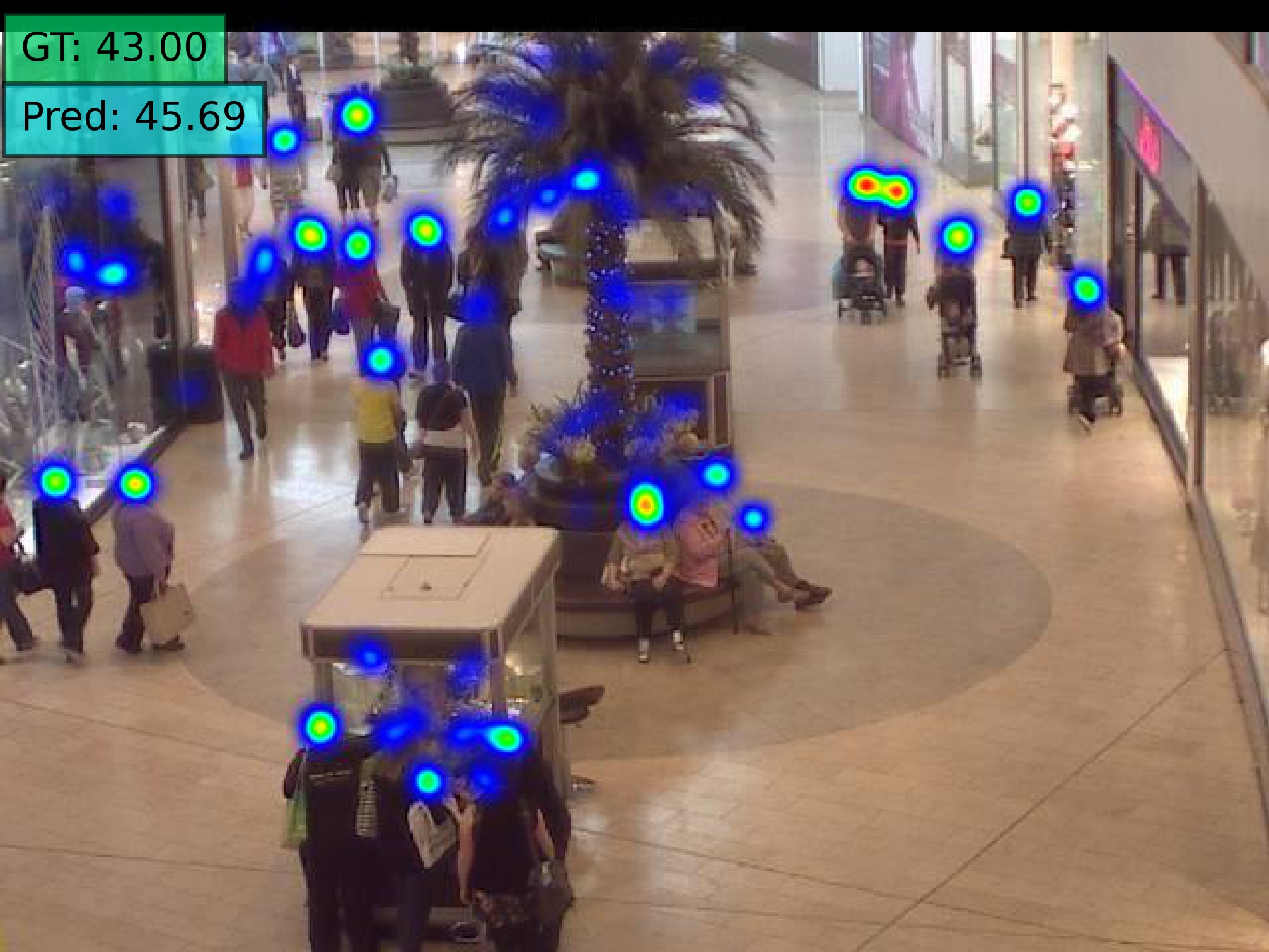}}&
          	\raisebox{-.5\height}{
          	\includegraphics[height=1.3in, width=1.4in]{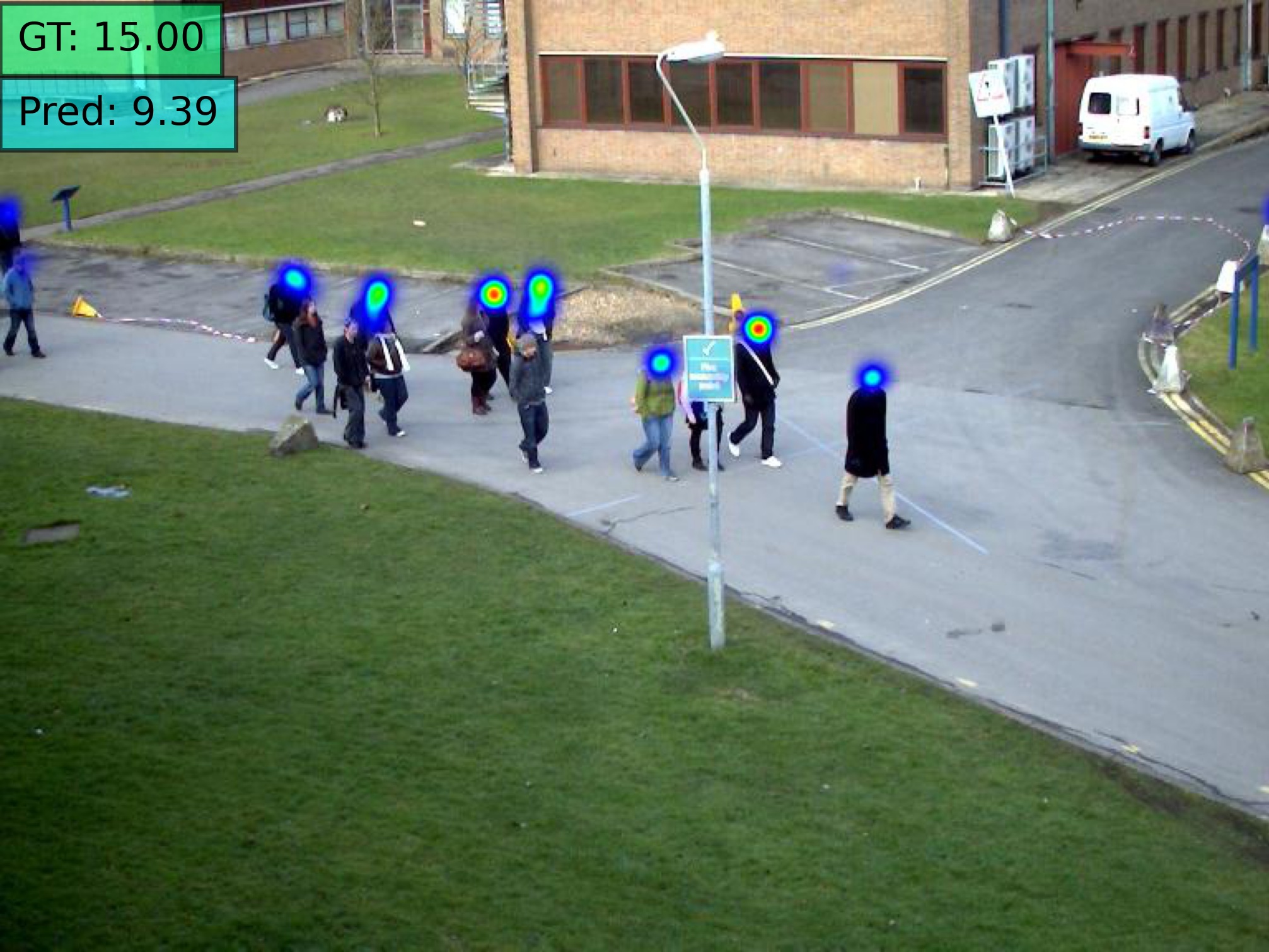}}&
          	\raisebox{-.5\height}{
          	\includegraphics[height=1.3in, width=1.4in]{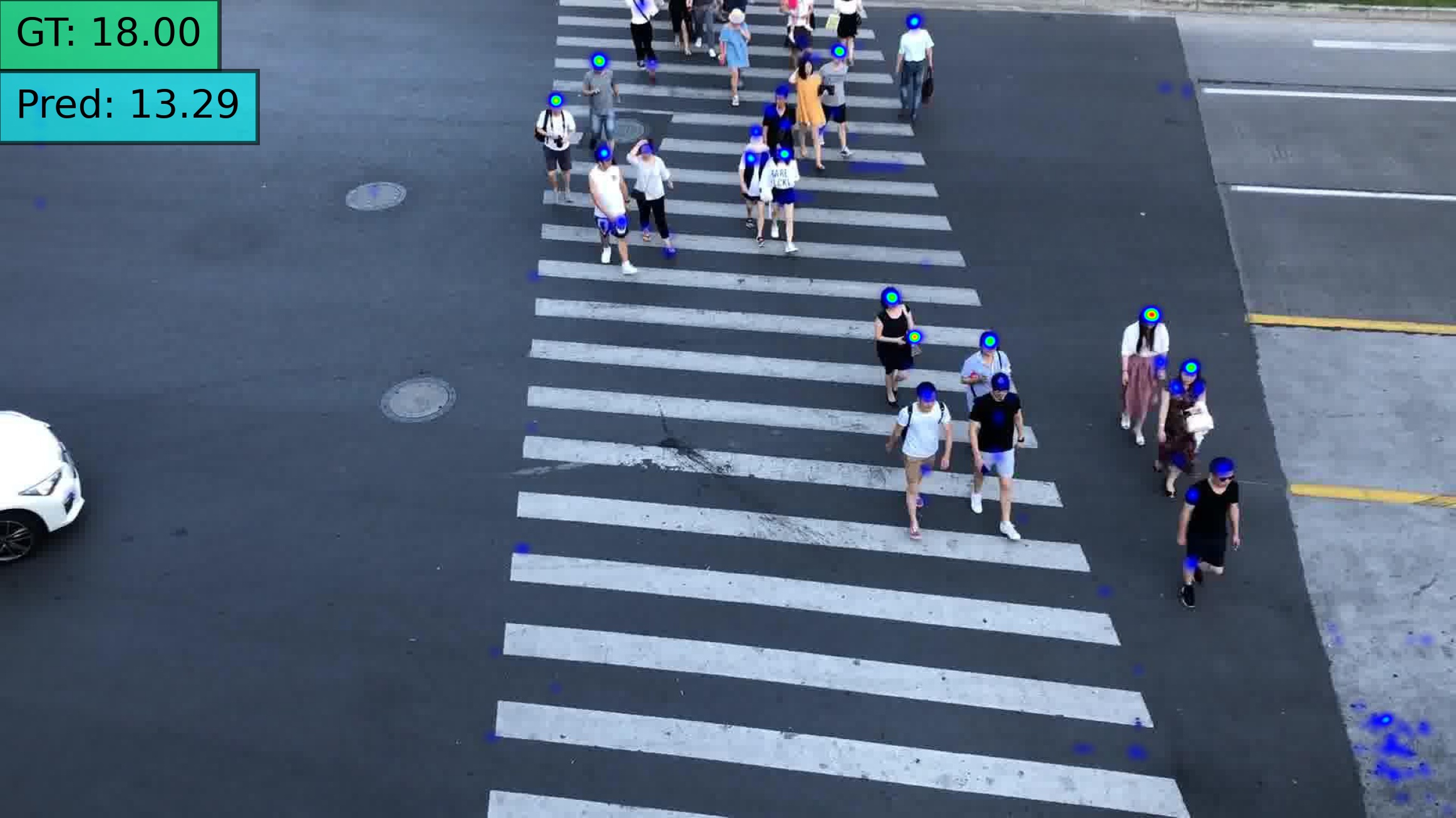}}\\
          	
		\textbf{Ours w/ CSRNet} &
		\raisebox{-.5\height}{
          	\includegraphics[height=1.3in, width=1.4in]{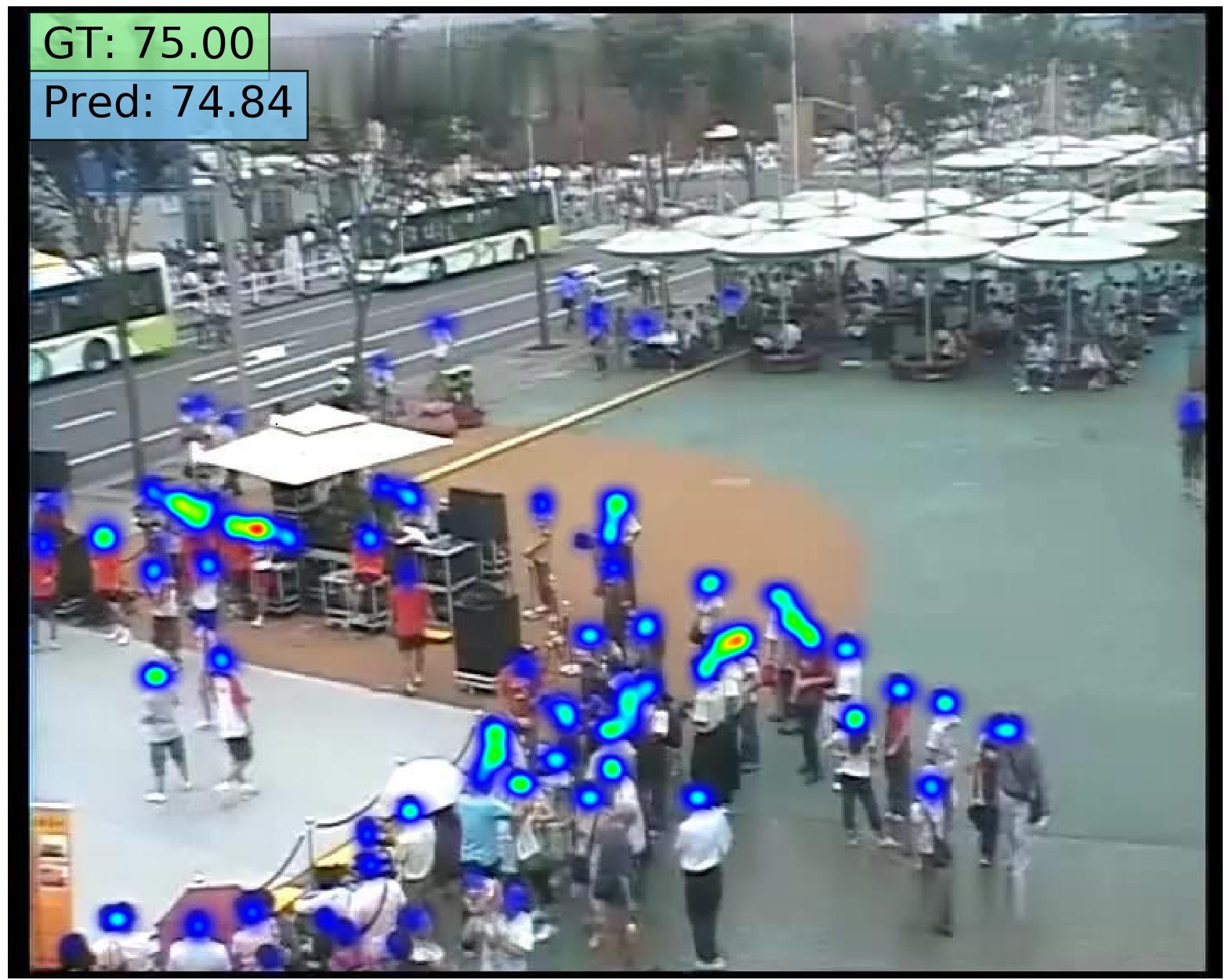}}&
          	\raisebox{-.5\height}{
          	\includegraphics[height=1.3in, width=1.4in]{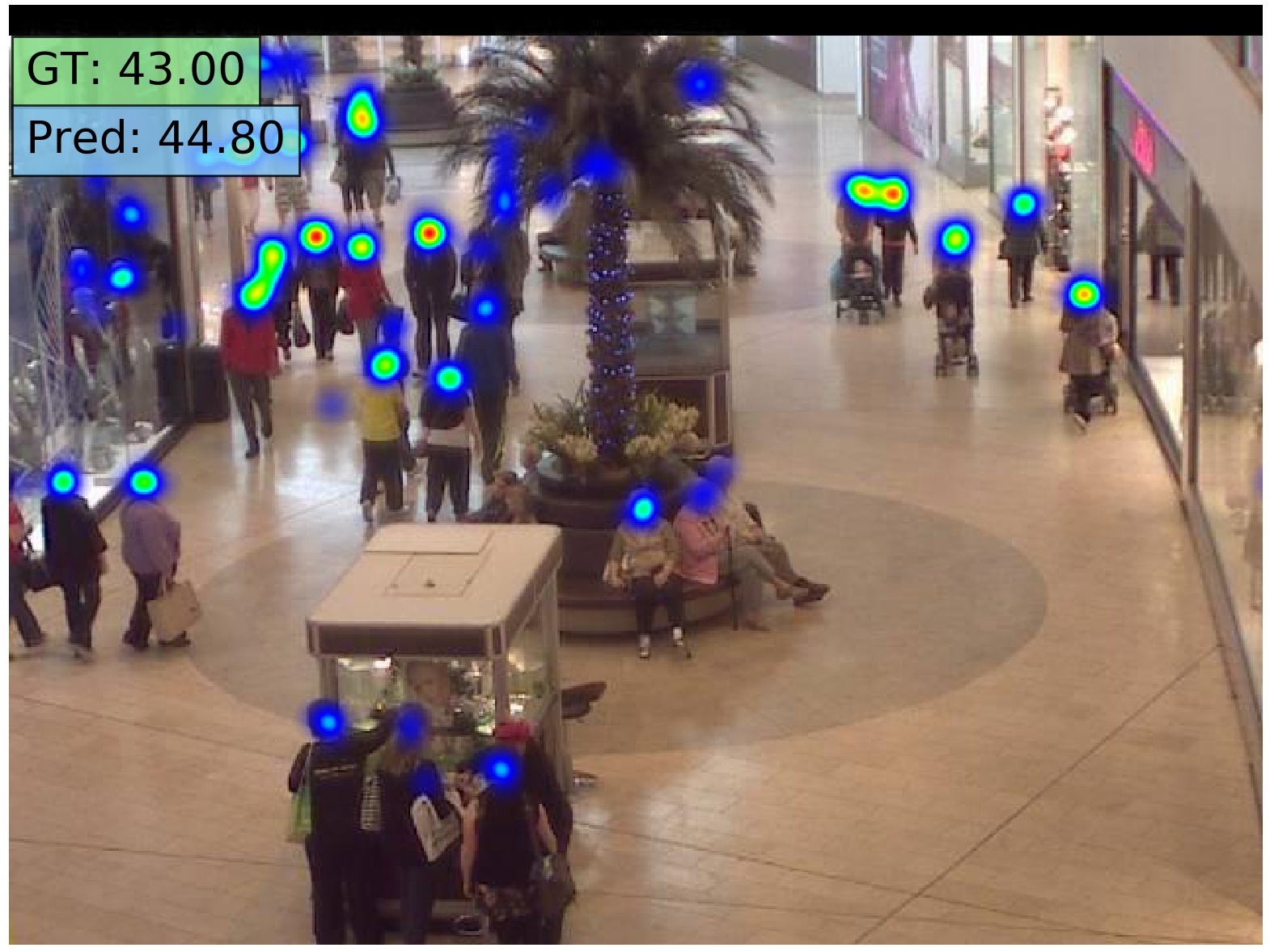}}&
          	\raisebox{-.5\height}{
          	\includegraphics[height=1.3in, width=1.4in]{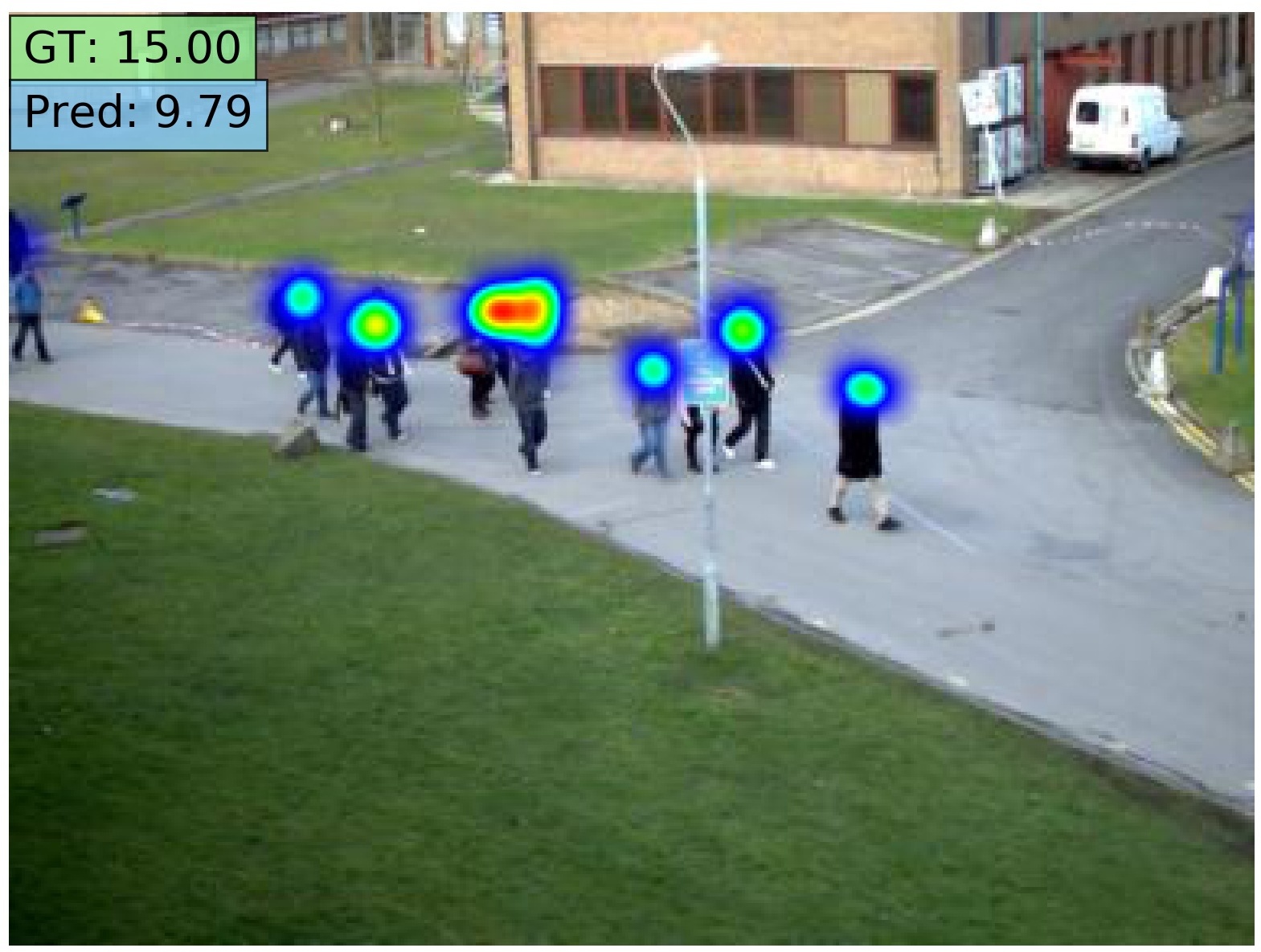}}&
          	\raisebox{-.5\height}{
          	\includegraphics[height=1.3in, width=1.4in]{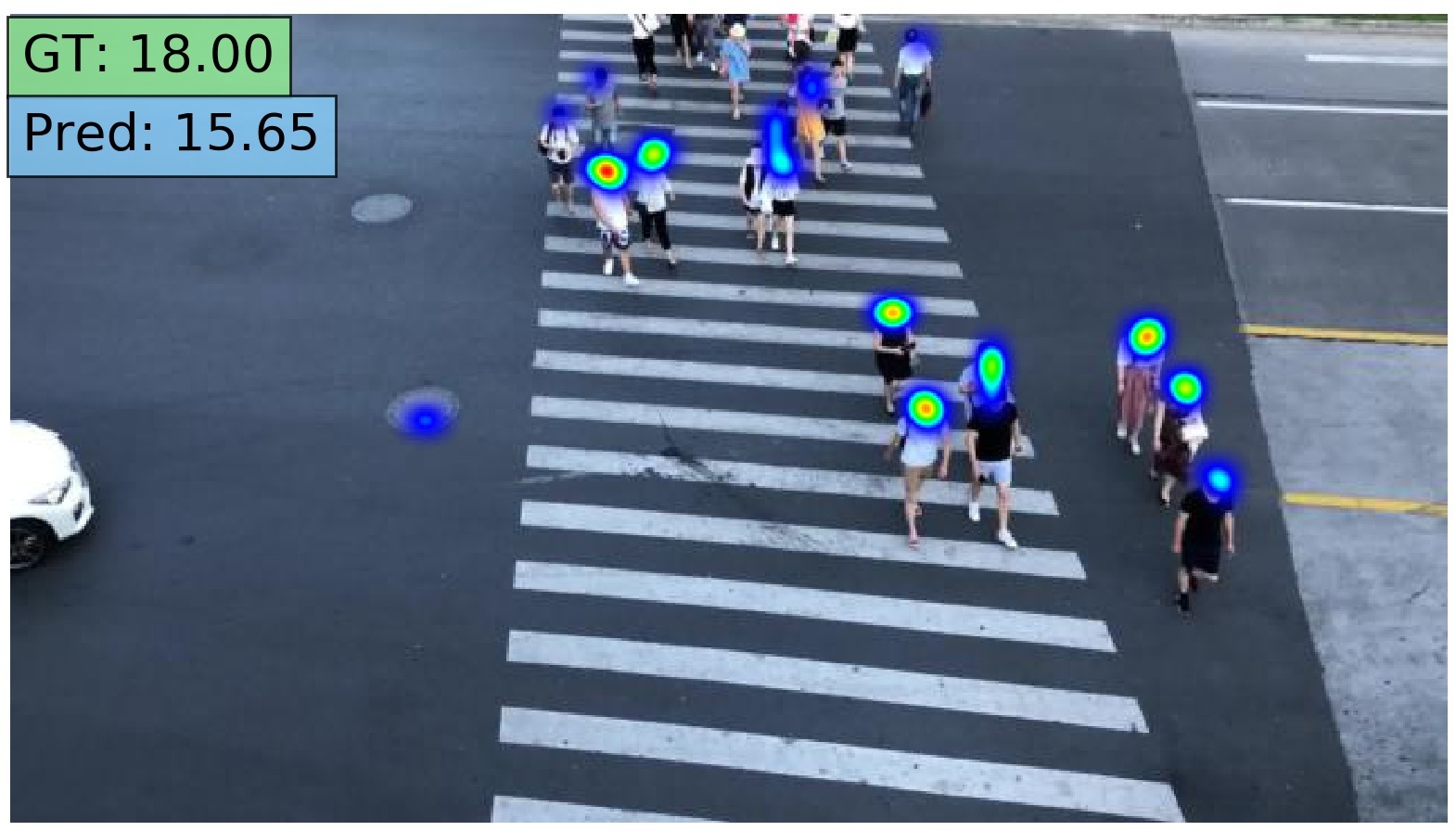}}\\

          & WorldExpo'10 & Mall & PETS & FDST\\
        \end{tabular}
        }
	\caption{Comparison of qualitative results between our proposed approach (\textbf{Ours w/ CSRNet}) and the baseline models (CSRNet~\cite{li2018csrnet} and CSRNet w/ BN ~\cite{li2018csrnet}) on different datasets. We visualize the predicted density map on each image. We also show both ground-truth and predicted counts at the top-left corner of each image.}
	\label{fig:qualitative}
\end{figure*}

\begin{table}
	\centering
        \caption{Quantitative results for one-unlabeled image based cross-dataset testing. We train all the models on WorldExpo'10 and test on the Venice dataset. We present the result for SFCN-based networks by reporting mean and standard deviation (\%) for our proposed approach. We show the best results in \textbf{bold}.}
	\renewcommand{\arraystretch}{1.2}
	\begin{tabular}{l|cc}
		\hline
		\multirow{2}{*}{Method} & \multicolumn{2}{c}{WorldExpo $\rightarrow$ Venice (1 input)} \\ \cline{2-3} 
		& MAE ($\downarrow$) & RMSE ($\downarrow$) \\
                \hline
                    SFCN & 147.48 & 157.48 \\ 
                  SFCN w/ BN  & 147.75 & 157.04 \\ 
                  \textbf{Ours w/ SFCN} & \textbf{133.83} \smaller{$\pm {\tiny 0.15}$}    & \textbf{146.44} \smaller{$\pm{\tiny  0.15}$}  \\ 
		\hline
	\end{tabular}
	\label{tab:we_venice_sfcn}
\end{table}

\begin{table*}
	\centering
        \caption{Comparison of our approach with 1 vs. 5 inputs (unlabeled image) by training and testing on WorldExpo'10. In all cases, the results of using 5 inputs are slightly better than using $1$ input. We report mean and standard deviation (\%) over 5 random trials for all the methods.}
	\renewcommand{\arraystretch}{1.2}
	\begin{tabular}{l|cc|cc}
		\hline
		\multirow{2}{*}{Method} & \multicolumn{2}{c}{1 input} & \multicolumn{2}{c}{5 inputs} \\ \cline{2-5} 
		& MAE ($\downarrow$) & RMSE ($\downarrow$) & MAE ($\downarrow$) & RMSE ($\downarrow$)\\
                \hline
                    Ours w/ CSRNet & 17.32 \smaller{$\pm {\tiny 0.1}$} & 27.03 \smaller{$\pm {\tiny 0.05}$} & 17.21 \smaller{$\pm {\tiny 0.6}$} & 26.85 \smaller{$\pm {\tiny 0.02}$} \\ 
                  Ours w/ FCN  & 20.91 \smaller{$\pm{\tiny 0.3}$}  & 29.61 \smaller{$\pm {\tiny 0.2}$} & 20.88 \smaller{$\pm {\tiny 0.2}$} & 29.53 \smaller{$\pm{\tiny 0.4}$} \\ 
                  Ours w/ SFCN & 14.56 \smaller{$\pm {\tiny 0.4}$}    & 22.75 \smaller{$\pm{\tiny 0.2}$} & 14.47 \smaller{$\pm{\tiny 0.4}$} & 22.61 \smaller{$\pm{\tiny 0.4}$} \\ 
		\hline
	\end{tabular}
	\label{tab:we_other_5shot}
\end{table*}

\begin{table}[ht]
	\centering
        \caption{Comparing the performance of two architecture setups using GBN layers on the WorldExpo'10 dataset.  In the first setup (marked by ``*''),  the GBN layers are used in both the encoder and decoder part of the crowd counting network.  In the second setup,  the GBN layers are used only in the decoder part of the crowd counting network. }
        \small
	\renewcommand{\arraystretch}{1}
	\begin{tabular}{ll|cc}
		\hline
		Backbone & Method	 & MAE ($\downarrow$)   & RMSE ($\downarrow$) \\ \hline
		VGG-16 & \texttt{Ours w/ CSRNet*}   & 48.46     & 59.84\\ 
		 & \textbf{Ours w/ CSRNet}   & \textbf{17.32}     & \textbf{27.03} \\ 
		 \hline
		 
		 ResNet-101		& \texttt{Ours w/ FCN*}   & 49.99     & 59.90\\ 
		 & \textbf{Ours w/ FCN}  & \textbf{20.91}    & \textbf{29.61} \\ 
		 \hline
		 
		 ResNet-101 & \texttt{Ours w/ SFCN*}   & 50.35     & 60.14\\ 
		 & \textbf{Ours w/ SFCN} & \textbf{14.56}    & \textbf{22.75} \\ 
			\hline
	\end{tabular}
    \label{tab:network_gbn_comparison}
\end{table}

\begin{table}[ht]
	\centering
        \caption{We compare our proposed unlabeled scene adaptation with one-shot~\cite{hossain2019one} and few-shot~\cite{Reddy_2020_WACV} adaptation methods for crowd counting on the WorldExpo'10 dataset.}
	\renewcommand{\arraystretch}{1.1}
	\begin{tabular}{l|cc}
		\hline
		Method	 & MAE ($\downarrow$)   & RMSE ($\downarrow$) \\ \hline
		 One-shot~\cite{hossain2019one}  & 8.23     & 12.08 \\
		 Few-shot~\cite{Reddy_2020_WACV}  &  7.5    & 10.22 \\
		 \textbf{Ours w/ CSRNet}   & 17.32     & 27.03 \\ 
		 \hline
	\end{tabular}
    \label{tab:compare_fewshot}
\end{table}

\subsection{Experimental Results}\label{sec:results}

\noindent \textbf{Quantitative Results:}
In Table~\ref{tab:we}, we show the average results of training with 103 scenes and testing on 5 different scenes on WorldExpo'10.  Note that since we do not use ROI maps, the results of CSRNet are slightly worse than the numbers reported in the original paper~\cite{li2018csrnet}.  In Table~\ref{tab:we_other}, we show the results of training on WorldExpo'10 and testing on other datasets (Mall, PETS, FDST, and CityUHK-X). Some of the datasets~(Mall, PETS, FDST) have the same scene in both training and testing. On these datasets, we use unlabeled images from the train set when sampling $z$ and evaluate the performance on all images from the corresponding test set. We perform 5 trials for each of our models on every dataset with different data on unlabeled image $z$. For the CityUHK-X dataset, the test scenes are different from train scenes. Therefore, we randomly select images from the test set for $z$ and evaluate on the remaining images from the test scenes. Similarly, we repeat this procedure 5 times. We finally report the mean score with the standard deviation (\%) in Table~\ref{tab:we} and~\ref{tab:we_other}.

Note that the CityUHK-X dataset is used in \cite{kang2017incorporating}. However, the work in \cite{kang2017incorporating} addresses a different problem from ours. In particular, the work in \cite{kang2017incorporating} exploits side-information (e.g.,  camera height, angle) to improve the crowd counting performance, while our work addresses unlabeled adaptation. We believe our problem setup is more applicable in the real world,  since the information about camera height/angle information is not always easily accessible in real-world applications. In fact, most crowd counting datasets do not have this information. In contrast, it is fairly easy to get an unlabeled image from a target scene. Due to the difference between the problem settings, a direct comparison with \cite{kang2017incorporating} is not possible since our method does not require or use the side-information on this dataset.  Nevertheless, we have evaluated our proposed method without using the side-information on the dataset from~\cite{kang2017incorporating} as shown in Table~\ref{tab:we_other}.  In all the cases, our {\fontfamily{qcr}\selectfont AdaCrowd} significantly outperforms other baselines.  

In Table~\ref{tab:we_venice_sfcn}, we present the results of training on WorldExpo'10 and evaluate for scene adaptation on the Venice~\cite{liu2019context} dataset. Similar to other quantitative results, we report mean and standard deviation for our proposed approach using SFCN-based networks over 5 random trials. Our proposed method performs better than the baselines.

For fair comparisons, we have used the same 5 random trials (e.g., the choice of unlabeled images) for the three backbone architectures in all the above experiments. So the improvements are only influenced by the choice of the backbone.

\noindent\textbf{Qualitative Results:} In Fig.~\ref{fig:qualitative}, we compare the qualitative examples between our method (\textbf{Ours w/ CSRNet}) and the baseline methods (CSRNet~\cite{li2018csrnet} and CSRNet w/ BN~\cite{li2018csrnet}) across four datasets. We visualize the predicted density map, the predicted count, and the ground-truth count. Our proposed approach consistently generates density maps closer to the ground-truth across different datasets. 

\begin{figure}[h]
		\centering
		\includegraphics[width=8.5cm]{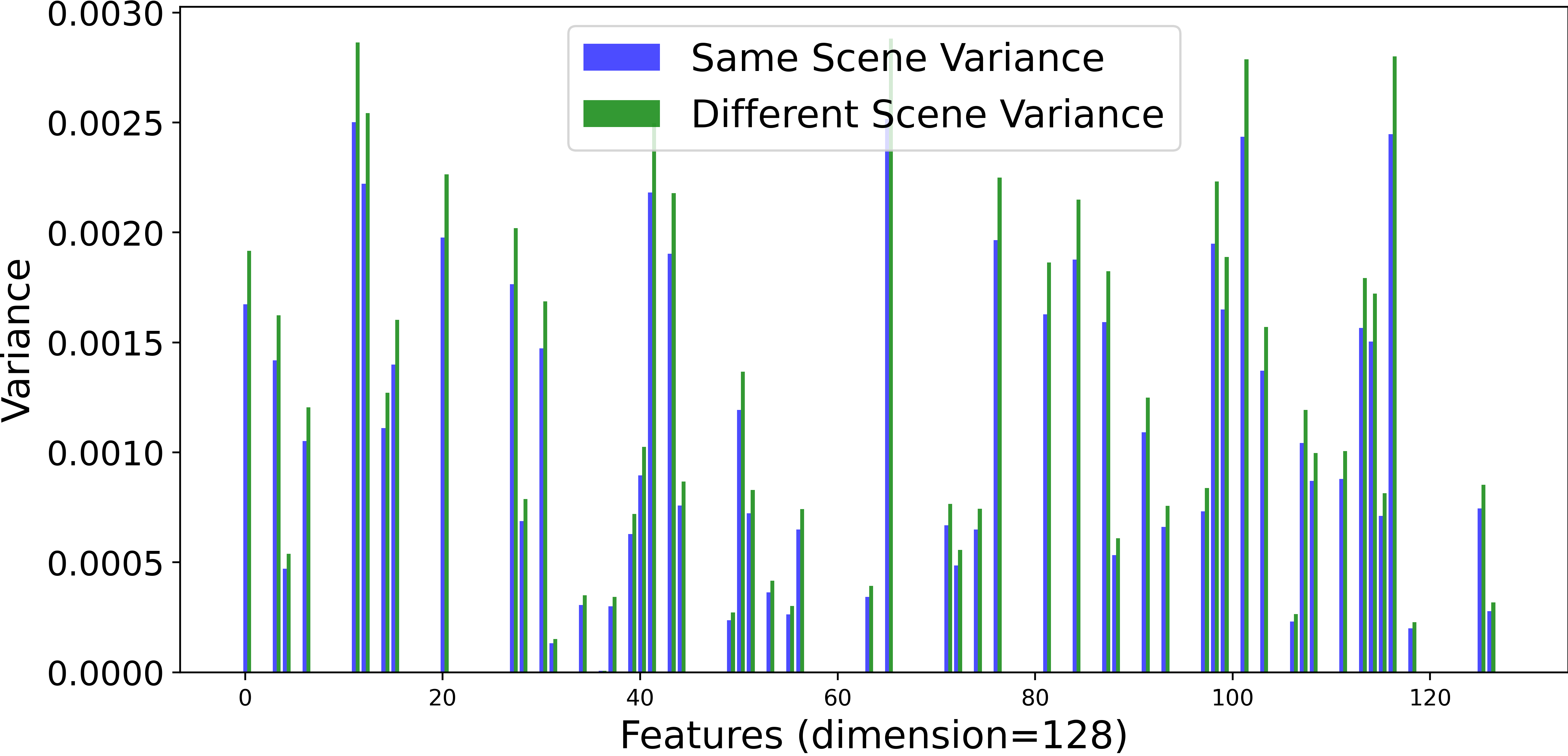}
	\caption{We plot the variance for every feature $\phi$ extracted from guiding network for the last GBN layer (with dimension 128) in the decoder on the WorldExpo'10 dataset.  (i) {\color{blue} \textbf{blue}} bars indicate the variance calculated over 10 images from the ``same'' scene; (ii) {\color{green} \textbf{green}} bars indicate the variance calculated over 10 images where each is sampled from a ``different'' scene.  In general, the variance for the features learned in the guiding network from the same scene is lower than that of images from different scenes.}
	\label{fig:std_dev}
\end{figure}

\begin{figure}[h]
		\centering
			\includegraphics[height=5cm]{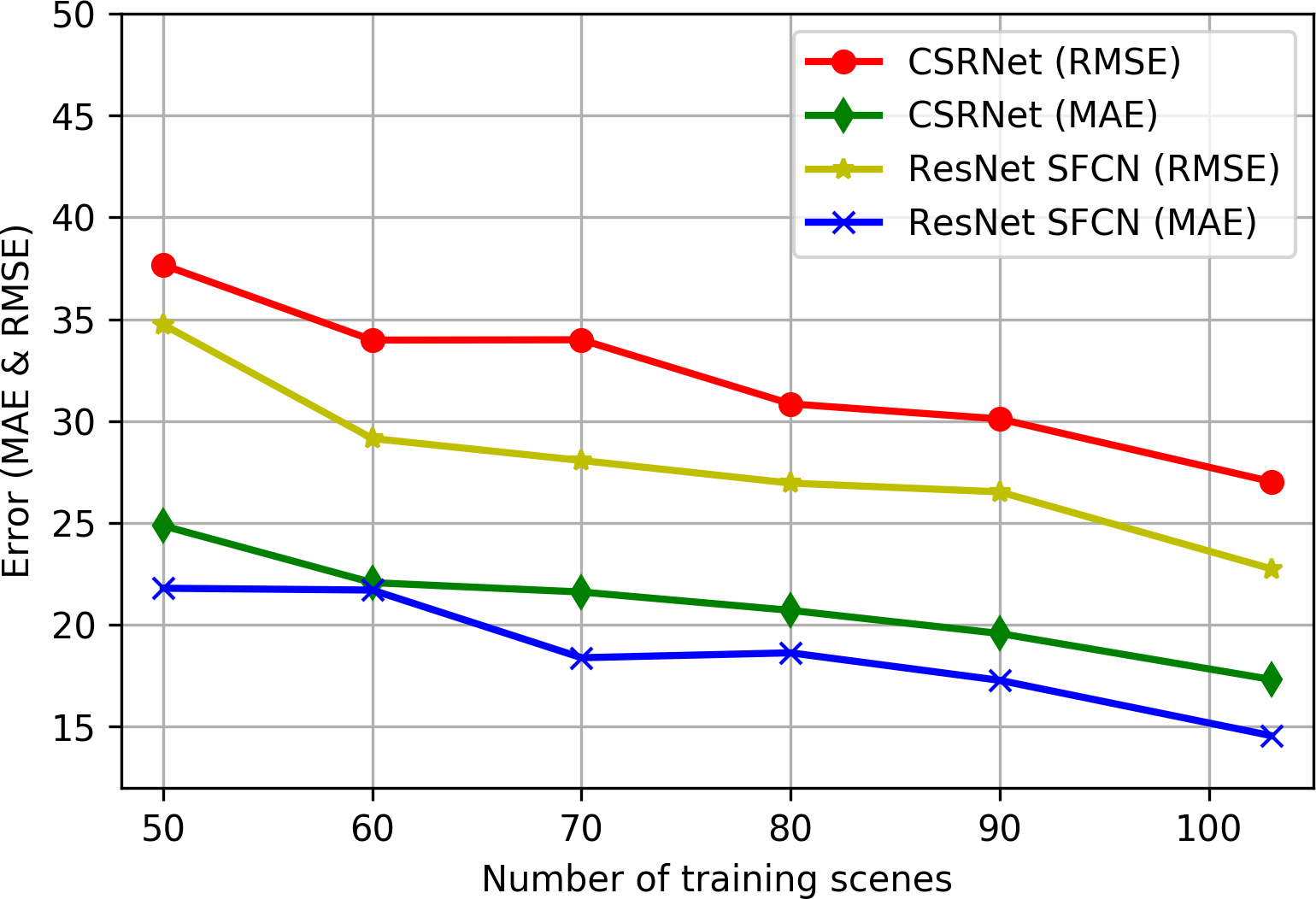}
		\caption{Performance vs. number of training scenes -- we present the ablation study on the relation between network performance and number of training scenes on the WorldExpo'10 dataset with CSRNet and ResNet SFCN networks.}
		\label{fig:scenes_vs_performance}
\end{figure}

\begin{figure}[h]
		\centering
			\includegraphics[height=5cm]{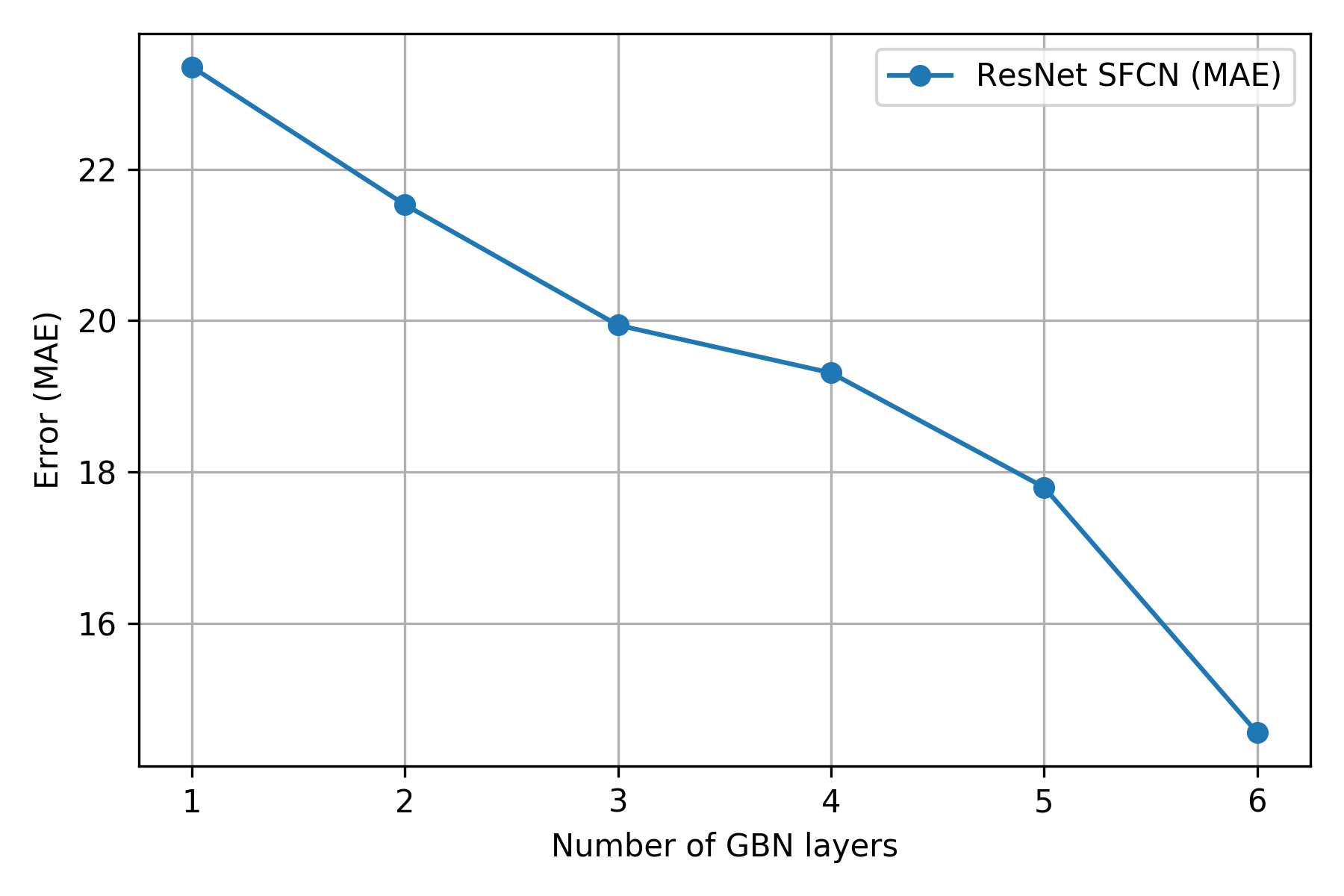}
		\caption{Performance vs. number of GBN layers -- an ablation study on the impact of the number of GBN layers on the network's performance when trained on the WorldExpo'10 dataset with ResNet SFCN architecture.}
		\label{fig:gbn_vs_performance}
\end{figure}

\begin{figure}[h]
		\centering
		\includegraphics[height=6cm]{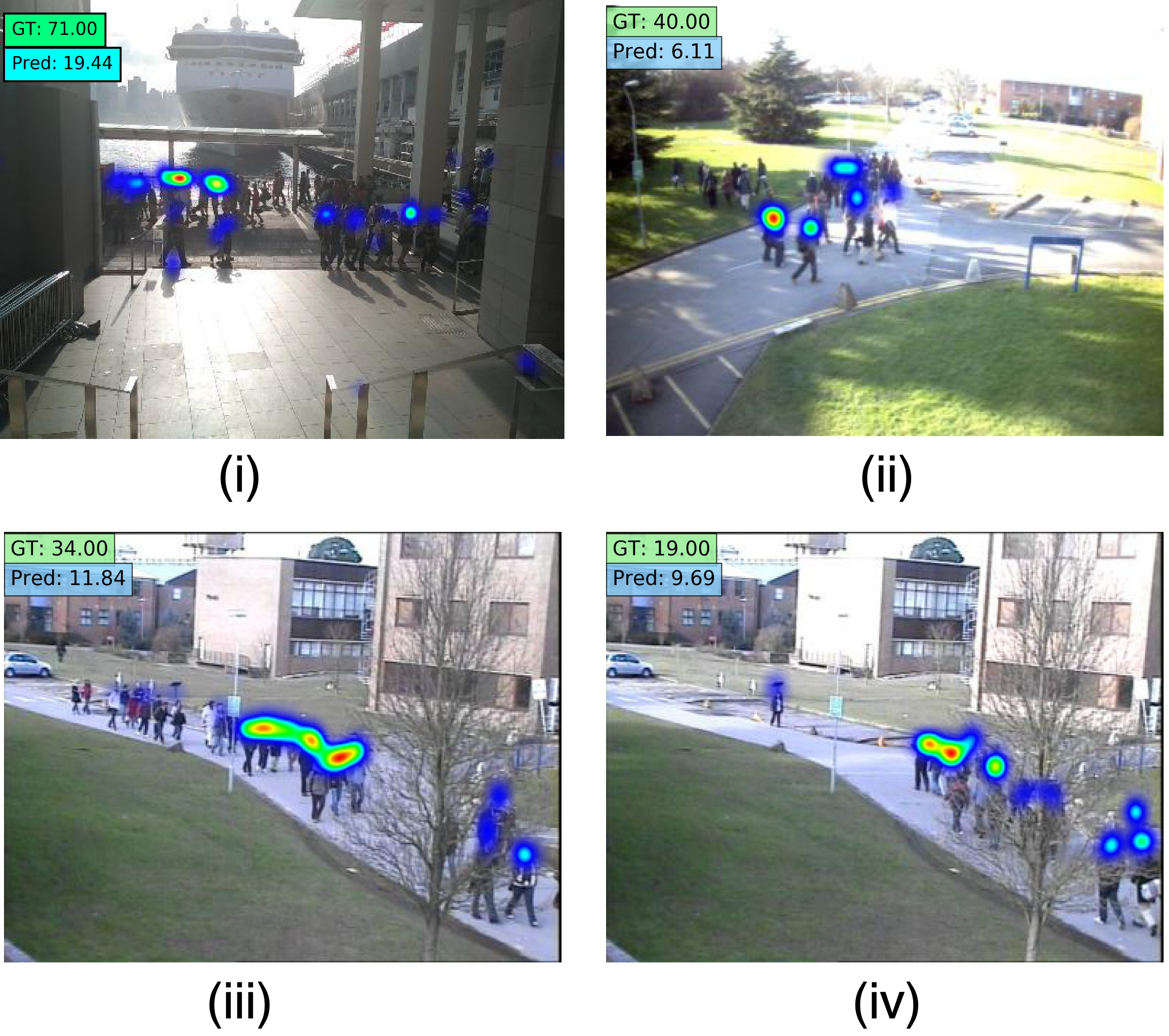}
	\caption{We provide an overview of some failure cases caused by drastic changes in the target scene images due to illumination, occlusion or image quality.  In the figure, (i) is a scene from the CityUHK-X dataset,  while (ii)-(iv) are scenes from the PETS dataset.  Nevertheless, our approach still performs better than alternative methods.}
	
	\label{fig:failure_cases}
\end{figure}

\noindent\textbf{Ablation Analysis:} We perform additional analysis on the proposed {\fontfamily{qcr}\selectfont AdaCrowd} framework to gain further understanding of the proposed approach. In this analysis, we increase the number of unlabeled images to 5. We show the results of training and testing on WorldExpo'10 in Table~\ref{tab:we_other_5shot}. In general, increasing the number of unlabeled data gives slightly better results. One possible explanation is that using more unlabeled data can make the algorithm more robust to noise. 

In Fig.~\ref{fig:architecture_overview}, we only use GBN layers in the decoder of the crowd counting network. The motivation for this design choice is as follows. The encoder of a crowd counting network typically uses existing backbone architectures (e.g., VGG,  FCN, SFCN) with pre-trained ImageNet weights. Some of these backbone architectures do not even contain batch normalization~(BN) layers. Even if they do, the pre-trained weights are coupled together with the BN parameters since they are trained jointly together. So it is difficult to insert GBN layers in the encoder without disrupting the information provided by the pre-trained weights. In contrast, the parameters in the decoder of the crowd counting network are initialized randomly, so the GBN layers can be learned together with other parameters in the decoder at the same time. In order to verify this, in Table~\ref{tab:network_gbn_comparison},  we compare different backbone architectures on two distinct configurations for the GBN layers on the WorldExpo'10 dataset: i) GBN layers used in both encoder and decoder; ii) GBN layers used only in the density map decoder.  The networks named ``Ours w/ CSRNet*'',  ``Ours w/ FCN*'' and ``Ours w/ SFCN*'' belong to the first category of using GBN layers in both the encoder and decoder part of the network. We can see that adding GBN layers in the encoder causes a significant drop in the performance. This is likely because these GBN layers in the encoder destroy the information provided by the pre-trained ImageNet weights used to initialize the encoder.

In Table~\ref{tab:compare_fewshot},  we compare our proposed setup for unlabeled scene adaptation with the previously proposed approaches for one-shot~\cite{hossain2019one} and few-shot~\cite{Reddy_2020_WACV} crowd scene adaptation to show the challenging nature of unlabeled scene adaptation. Not surprisingly, the performance of unlabeled scene adaptation is not as good as one-shot or few-shot settings, since the unlabeled scene adaptation uses much less information (only unlabeled images), while the settings in \cite{hossain2019one,Reddy_2020_WACV} require the end-user to provide manual annotations for some images from the target scene.

In Fig.~\ref{fig:std_dev},  we visualize the variance of the feature $\phi$ extracted for 10 images from the WorldExpo'10 dataset by the guiding network for two data setups for the last GBN layer in the decoder. The two input data scenarios are: i) different images from the same scene; ii) different images from diverse scenes.  In the former case, we randomly sample 10 different images from the same scene and extract their corresponding features from the guiding network. We then plot the feature-wise variance for all images. Similarly, in the latter case,  we randomly sample 10 images from 10 different scenes and plot their feature-wise variance. To make the figure less cluttered, we only show the visualization of $\phi$ corresponding to the last GBN layer (with dimension 128). From the figure, it is clear that the variance within a scene is lower than the variance across different scenes for the features extracted by the guiding network. This indicates that the network shows less randomness towards images from the same scene while showing more randomness in the features extracted from different scenes. This confirms that the guiding network extracts scene-variant features.

In Fig.~\ref{fig:scenes_vs_performance}, we present the analysis on how the performance varies when we vary the number of training scenes. This analysis uses the CSRNet and ResNet SFCN architectures on the WorldExpo'10 dataset. The performance (low error for MAE and RMSE) is positively correlated with the number of training scenes. Therefore, {\fontfamily{qcr}\selectfont AdaCrowd} framework performs better at test time if more scenes are available during training.

In Fig.~\ref{fig:gbn_vs_performance}, we study the performance of the SFCN-based network by varying the number of GBN layers in the second sub-network (decoder) of the framework that generates the output density map. Interestingly, increasing the GBN layers to 6 yields the best performance on the WorldExpo'10 dataset. This is also equivalent to adding a GBN layer for every dilated convolutional layer in the second sub-network.

In Fig.~\ref{fig:failure_cases},  we show some failure cases due to factors such as illumination, occlusion or image quality that have drastic effect in the target scene images. However, our approach still performs better than other methods in these cases. As future work, we will explore how to incorporate other meta-information (e.g., illumination estimation) into our framework to deal with these challenging cases.
\section{Conclusion}
We have introduced a new problem called the unlabeled scene-adaptive crowd counting. Our goal is to adapt a crowd counting model to a target scene using some unlabeled data from that scene. Compared with existing problem setups, this new problem setup is closer to the real-world deployment of crowd counting systems. We have proposed a novel framework {\fontfamily{qcr}\selectfont AdaCrowd} with GBN layers for solving this problem. Our proposed approach employs a guiding network to predict GBN parameters of the crowd counting network based on the unlabeled data of a scene. The model parameters are learned in a way that allows effective adaptation to new scenes given their unlabeled data. Our experimental results demonstrate that our proposed approach outperforms other alternative methods.


\bibliographystyle{IEEEtran}
\bibliography{IEEEabrv, mahesh}
%

%

%
%
%




\section{Supplementary Material}
\subsection{Network Architecture}

In this section, we describe the different network components employed in our framework.  In case of the backbone crowd counting network (i.e., encoder), we experimented on both VGG-16~\cite{simonyan2014very} and ResNet-101~\cite{he2016deep} architectures.  Similarly for the density map generator (i.e., decoder), we used the a series of dilated convolution layers as proposed in CSRNet~\cite{li2018csrnet} for our CSRNet and ResNet FCN networks.  Our ResNet SFCN decoder follows the architecture proposed in~\cite{wang2019learning} that is similar to the decoder in CSRNet,  but with more additional spatial layers.

\begin{table}[!h]
\centering
\caption{The network architecture for CSRNet~\cite{li2018csrnet} encoder uses VGG-16~\cite{simonyan2014very} as the backend network pretrained on ImageNet~\cite{deng2009imagenet}.  In our experiments,  we use VGG-16 upto $\texttt{conv\_4\_3}$.  In the table,  ``$k$'' indicates the kernel size and ``$s$'' is the filter stride.}
\begin{tabular}{lll}
\hline
\multicolumn{1}{c}{\textbf{Layer}}  & \multicolumn{1}{c}{\textbf{Configuration}} & \multicolumn{1}{c}{\textbf{Filter}}             \\ \cline{1-3}
\multicolumn{1}{c}{Conv2d} & \multicolumn{1}{c}{$k$(3,3),  $s$1}  & \multicolumn{1}{c}{3 $\rightarrow$ 64} \\ 
\multicolumn{3}{c}{ReLU}                                                                                                                                              \\ 
\multicolumn{1}{c}{Conv2d} & \multicolumn{1}{c}{$k$(3,3),  $s$1}  & \multicolumn{1}{c}{64 $\rightarrow$ 64} \\ 
\multicolumn{3}{c}{ReLU}                                                                                                                                         \\
\multicolumn{1}{c}{MaxPool2d} & \multicolumn{1}{c}{$k$(2,2),  $s$2}  &  \\ 
 \hline
\multicolumn{1}{c}{Conv2d} & \multicolumn{1}{c}{$k$(3,3),  $s$1}  & \multicolumn{1}{c}{64 $\rightarrow$ 128} \\ 
\multicolumn{3}{c}{ReLU}                                                                                                                                            \\ 
\multicolumn{1}{c}{Conv2d} & \multicolumn{1}{c}{$k$(3,3),  $s$1}  & \multicolumn{1}{c}{128 $\rightarrow$ 128} \\ 
\multicolumn{3}{c}{ReLU}                                                                                                                                            \\ 
\multicolumn{1}{c}{MaxPool2d} & \multicolumn{1}{c}{$k$(2,2),  $s$2}  &  \\ 
 \hline
\multicolumn{1}{c}{Conv2d} & \multicolumn{1}{c}{$k$(3,3),  $s$1}  & \multicolumn{1}{c}{128 $\rightarrow$ 256} \\ 
\multicolumn{3}{c}{ReLU}                                                                                                                                            \\ 
\multicolumn{1}{c}{Conv2d} & \multicolumn{1}{c}{$k$(3,3),  $s$1}  & \multicolumn{1}{c}{256 $\rightarrow$ 256} \\ 
\multicolumn{3}{c}{ReLU}                                                                                                                                            \\ 
\multicolumn{1}{c}{Conv2d} & \multicolumn{1}{c}{$k$(3,3),  $s$1}  & \multicolumn{1}{c}{256 $\rightarrow$ 256} \\ 
\multicolumn{3}{c}{ReLU}                                                                                                                                             \\ 
\multicolumn{1}{c}{MaxPool2d} & \multicolumn{1}{c}{$k$(2,2),  $s$2}  &  \\ 
 \hline
\multicolumn{1}{c}{Conv2d} & \multicolumn{1}{c}{$k$(3,3),  $s$1}  & \multicolumn{1}{c}{256 $\rightarrow$ 512} \\ 
\multicolumn{3}{c}{ReLU}                                                                                                                                            \\ 
\multicolumn{1}{c}{Conv2d} & \multicolumn{1}{c}{$k$(3,3),  $s$1}  & \multicolumn{1}{c}{512 $\rightarrow$ 512} \\ 
\multicolumn{3}{c}{ReLU}                                                                                                                                            \\ 
\multicolumn{1}{c}{Conv2d} & \multicolumn{1}{c}{$k$(3,3),  $s$1}  & \multicolumn{1}{c}{512 $\rightarrow$ 512} \\ 
\multicolumn{3}{c}{ReLU}                                                                                                                                             \\ 
 \hline
\end{tabular}
\end{table}

\begin{table}[!t]
\centering
\caption{The network architecture for ResNet FCN and SFCN~\cite{wang2019learning} encoders use ResNet-101~\cite{he2016deep} as the backend network pretrained on ImageNet~\cite{deng2009imagenet}.  In our experiments,  we use VGG-16 upto \textit{3rd} ResNet block.  In the table,  ``$k$'' indicates the kernel size and ``$s$'' is the filter stride.}
\begin{tabular}{lll}
\hline
\multicolumn{1}{c}{\textbf{Layer}}  & \multicolumn{1}{c}{\textbf{Configuration}} & \multicolumn{1}{c}{\textbf{Filter}}             \\ \cline{1-3}
\multicolumn{1}{c}{Conv2d} & \multicolumn{1}{c}{$k$(7,7),  $s$1}  & \multicolumn{1}{c}{3 $\rightarrow$ 64} \\ 
\multicolumn{3}{c}{BatchNorm2d(64)}                                                                                                                                              \\ 
\multicolumn{3}{c}{ReLU}                                                                                                                                              \\ 
\multicolumn{1}{c}{MaxPool2d} & \multicolumn{1}{c}{$k$(3,3),  $s$2}  &  \\ 
 \hline
 \multirow{3}{*}{ResBlock1}       & \multirow{1}{*}{
$\begin{bmatrix}    
k(1,1),  64\\
k(3,3),  64\\
k(1,1),  256\\
        \end{bmatrix} \times 3$}          & \multirow{3}{*}{}  \\
                            &                             &   \\
                                                        &                             &   \\
                                                         \hline
 \multirow{3}{*}{ResBlock2}       & \multirow{1}{*}{
$\begin{bmatrix}    
k(1,1), 128\\
k(3,3),  128\\
k(1,1), 512\\
        \end{bmatrix} \times 4$}          & \multirow{3}{*}{}  \\
                            &                             &   \\
                                                        &                             &   \\
                                                        \hline
 \multirow{3}{*}{ResBlock3}       & \multirow{1}{*}{
$\begin{bmatrix}    
k(1,1),  256\\
k(3,3),  256\\
k(1,1), 1024\\
        \end{bmatrix} \times 23$}          & \multirow{3}{*}{}  \\
                            &                             &   \\
                                                        &                             &   \\
\hline
\end{tabular}
\end{table}

\begin{table}[]
\centering
\caption{The network architecture of guiding network to map from an unlabeled image $z$ to the GBN parameters $\phi$.  The network consists of three convolution layers with ReLU activation,  an AdaptiveAveragePool2D layer to reduce the spatial feature dimensions and a linear layer to generate 3968 GBN parameters.  In the table,  ``$k$'' indicates the kernel size, ``$s$'' is the filter stride and padding ``$p$'' is the padding.}
\begin{tabular}{lll}
\hline
\multicolumn{1}{c}{\textbf{Layer}}  & \multicolumn{1}{c}{\textbf{Configuration}} & \multicolumn{1}{c}{\textbf{Filter}}             \\ \cline{1-3}
\multicolumn{1}{c}{Conv2d} & \multicolumn{1}{c}{$k$(7,7),  $s$1,  $p$3}  & \multicolumn{1}{c}{3 $\rightarrow$ 64} \\ 
\multicolumn{3}{c}{ReLU}                                                                                                                                              \\ 
\multicolumn{1}{c}{Conv2d} & \multicolumn{1}{c}{$k$(4,4),  $s$2,  $p$1}  & \multicolumn{1}{c}{64 $\rightarrow$ 128} \\ 
\multicolumn{3}{c}{ReLU}                                                                                                                                         \\
\multicolumn{1}{c}{Conv2d} & \multicolumn{1}{c}{$k$(4,4),  $s$2,  $p$1}  & \multicolumn{1}{c}{128 $\rightarrow$ 256} \\ 
\multicolumn{3}{c}{ReLU} \\                                                                                                                                         
\multicolumn{3}{c}{AdaptiveAveragePool2d(1x1)}                                                                                                                                         \\
\multicolumn{1}{c}{Linear} & \multicolumn{1}{c}{}  & \multicolumn{1}{c}{256 $\rightarrow$ 3968} \\ 
 \hline
\end{tabular}
\end{table}

\begin{table}[!h]
\centering
\caption{The network architecture for the crowd counting density map generator (i.e.,  decoder) is based on~\cite{li2018csrnet}.  We have incorporated our proposed GuidedBatchNorm2d (GBN) layers to module the input features from the crowd counting encoder network.  The six GBN layers consists of a total of 3968 parameters that are predicted from the guiding network.  In the table,  ``$k$'' indicates the kernel size,``$s$'' is the filter stride,  ``$p$'' is the padding and ``$d$'' is the dilation rate.}
\begin{tabular}{lll}
\hline
\multicolumn{1}{c}{\textbf{Layer}}  & \multicolumn{1}{c}{\textbf{Configuration}} & \multicolumn{1}{c}{\textbf{Filter}}             \\ \cline{1-3}
\multicolumn{1}{c}{Conv2d} & \multicolumn{1}{c}{$k$(3,3),  $s$1,  $p$2,  $d$2}  & \multicolumn{1}{c}{(1024/512) $\rightarrow$ 512} \\ 
\multicolumn{3}{c}{GuidedBatchNorm2d(512)}                                                                                                                                              \\ 
\multicolumn{3}{c}{ReLU}                                                                                                                                              \\ 
 \multicolumn{1}{c}{Conv2d} & \multicolumn{1}{c}{$k$(3,3),  $s$1,  $p$2,  $d$2}  & \multicolumn{1}{c}{512 $\rightarrow$ 512} \\ 
\multicolumn{3}{c}{GuidedBatchNorm2d(512)}                                                                                                                                              \\ 
\multicolumn{3}{c}{ReLU}                                                                                                                                              \\ 
 \multicolumn{1}{c}{Conv2d} & \multicolumn{1}{c}{$k$(3,3),  $s$1,  $p$2,  $d$2}  & \multicolumn{1}{c}{512 $\rightarrow$ 512} \\ 
\multicolumn{3}{c}{GuidedBatchNorm2d(512)}                                                                                                                                              \\ 
\multicolumn{3}{c}{ReLU}                                                                                                                                              \\ 
 \multicolumn{1}{c}{Conv2d} & \multicolumn{1}{c}{$k$(3,3),  $s$1,  $p$2,  $d$2}  & \multicolumn{1}{c}{512 $\rightarrow$ 256} \\ 
\multicolumn{3}{c}{GuidedBatchNorm2d(256)}                                                                                                                                              \\ 
\multicolumn{3}{c}{ReLU}                                                                                                                                              \\ 
\hline
 \multicolumn{1}{c}{Conv2d} & \multicolumn{1}{c}{$k$(3, 3),  $s$1,  $p$2,  $d$2}  & \multicolumn{1}{c}{256 $\rightarrow$ 128} \\ 
\multicolumn{3}{c}{GuidedBatchNorm2d(128)}                                                                                                                                              \\ 
\multicolumn{3}{c}{ReLU}                                                                                                                                              \\ 
\hline
 \multicolumn{1}{c}{Conv2d} & \multicolumn{1}{c}{$k$(3,3),  $s$1,  $p$2,  $d$2}  & \multicolumn{1}{c}{128 $\rightarrow$ 64} \\ 
\multicolumn{3}{c}{GuidedBatchNorm2d(64)}                                                                                                                                              \\ 
\multicolumn{3}{c}{ReLU}                                                                                                                                              \\ 
\hline
 \multicolumn{3}{c}{\textbf{This block is used only for ResNet SFCN architecture}} \\ 
  \multicolumn{3}{c}{} \\ 
 \multicolumn{1}{c}{Conv2d} & \multicolumn{1}{c}{$k$(1, 9),  $s$1,  $p$(0,4),  $d$1 }  & \multicolumn{1}{c}{64 $\rightarrow$ 64} \\ 
 \multicolumn{3}{c}{ReLU}                                                                                                                                              \\ 
  \multicolumn{1}{c}{Conv2d} & \multicolumn{1}{c}{$k$(9,1),  $s$1,  $p$(4,0),  $d$1}  & \multicolumn{1}{c}{64 $\rightarrow$ 64} \\ 
 \multicolumn{3}{c}{ReLU}                                                                                                                                              \\ 
\hline
 \multicolumn{1}{c}{Conv2d} & \multicolumn{1}{c}{$k$(1,1),  $s$1,  $p$1,  $d$1}  & \multicolumn{1}{c}{64 $\rightarrow$ 1} \\ 
 \hline
\end{tabular}
\end{table}

\end{document}